\documentclass[10pt,twocolumn,letterpaper]{article}

\usepackage{cvpr}              % To produce the CAMERA-READY version
\usepackage[dvipsnames]{xcolor}

\newcommand{\arch}[1]{\textsc{#1}}

\definecolor{cvprblue}{rgb}{0.21,0.49,0.74}
\usepackage[pagebackref,breaklinks,colorlinks,citecolor=cvprblue]{hyperref}

\newcommand\blfootnote[1]{%
  \begingroup
  \renewcommand\thefootnote{}\footnote{#1}%
  \addtocounter{footnote}{-1}%
  \endgroup
}

\def\paperID{11668} % *** Enter the Paper ID here
\def\confName{CVPR}
\def\confYear{2024}

\title{Style Injection in Diffusion: A Training-free Approach\\ for Adapting Large-scale Diffusion Models for Style Transfer}

\author{Jiwoo Chung$^{\ast}$, Sangeek Hyun$^{\ast}$, Jae-Pil Heo$^{\dag}$ \\
Sungkyunkwan University\\
{\tt\small {\{wldn0202, hsi1032, jaepilheo\}@g.skku.edu}}}

\newcommand{\Skip}[1]{}

\begin{document}
\maketitle
\begin{abstract}
Despite the impressive generative capabilities of diffusion models, existing diffusion model-based style transfer methods require inference-stage optimization~(e.g. fine-tuning or textual inversion of style) which is time-consuming, or fails to leverage the generative ability of large-scale diffusion models.
To address these issues, we introduce a novel artistic style transfer method based on a pre-trained large-scale diffusion model without any optimization.
Specifically, we manipulate the features of self-attention layers as the way the cross-attention mechanism works; 
in the generation process, substituting the key and value of content with those of style image.
This approach provides several desirable characteristics for style transfer including 1) preservation of content by transferring similar styles into similar image patches and 2) transfer of style based on similarity of local texture~(e.g. edge) between content and style images.
Furthermore, we introduce query preservation and attention temperature scaling to mitigate the issue of disruption of original content, and initial latent Adaptive Instance Normalization~(AdaIN) to deal with the disharmonious color~(failure to transfer the colors of style).
Our experimental results demonstrate that our proposed method surpasses state-of-the-art methods in both conventional and diffusion-based style transfer baselines. Codes are available at \href{https://github.com/jiwoogit/StyleID}{github.com/jiwoogit/StyleID}.

\blfootnote{
    $^\ast$ Equal contribution
}
\blfootnote{
$^\dag$ Corresponding author
}
\end{abstract}
    
\vspace{-0.2cm}
\section{Introduction}
\label{sec:intro}
Recent advances in Diffusion Models~(DMs) have led to breakthroughs in various generative applications such as text-to-image synthesis~\cite{rombach2022high,nichol2021glide,saharia2022photorealistic} and image or video editing~\cite{couairon2023diffedit,kawar2023imagic,tumanyan2023plug,hertz2022prompt,avrahami2022blended,chai2023stablevideo,yang2023rerender}.
One of these efforts is also applied to the task of style transfer~\cite{jeong2023training,wang2023stylediffusion,everaert2023diffusion,yang2023zero,zhang2023inversion}; given style and content images, modifying the style of the content image to possess the given style.

\begin{figure}[t!]
    \includegraphics[width=1.0\columnwidth]{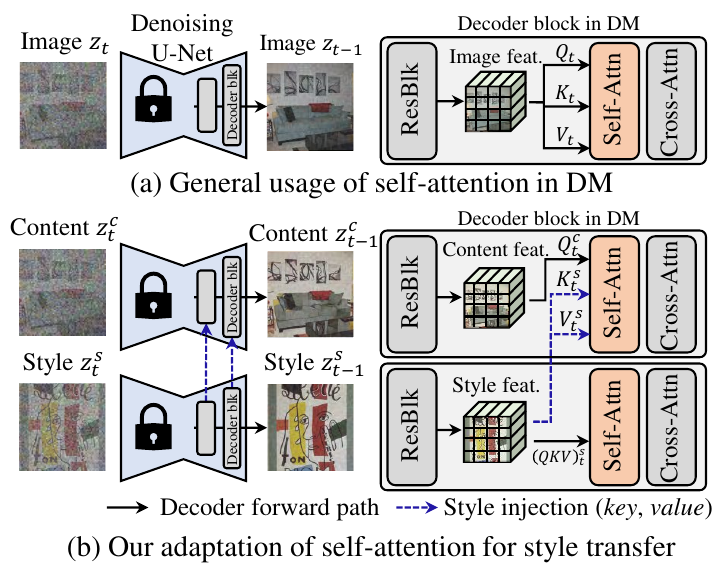}
    \centering
    \vspace{-0.5cm}
    \caption{
    \textbf{Manipulation of self-attention features for style transfer.} 
    (a) General self-attention~(SA) deploys the \textit{query}, \textit{key}, and \textit{value} features from a single image in both the training and inference phases.
    (b) At inference phase, we suggest that manipulating features of self-attention of pre-trained large-scale DM is an effective way to transfer the styles; injection of \textit{key} and \textit{value} of styles into SA of contents is a proper way for transferring styles. 
    As a result, style-injected content $z^c_{t-1}$ would maintain contents while modifying its style to resemble the target style. 
    }
    \vspace{-0.5cm}
    \label{fig_motivation_self_attn_method}
\end{figure}

General approaches for diffusion model-based style transfer leverage the generative capability of pre-trained DM.
Some of these works focus on explicit disentangling style and content for interpretable and controllable style transfer~\cite{wang2023stylediffusion}, or inversion of the style image into the textual latent space of a large-scale text-to-image DM~\cite{zhang2023inversion}.
However, these methods additionally require gradient-based optimization for fine-tuning and textual inversion~\cite{ruiz2023dreambooth} for each style image, which is time-consuming. 
Without this issue, DiffStyle~\cite{jeong2023training} introduces training-free style transfer,
but they are known to be hardly applicable to Latent Diffusion Model~\cite{rombach2022high} which is widely adopted for training large-scale text-to-image DM such as Stable Diffusion~\cite{rombach2022high}, hindering the users from taking advantage of the prominent generative ability of large-scale models.

\begin{figure}[t!]
    \includegraphics[width=1.0\columnwidth]{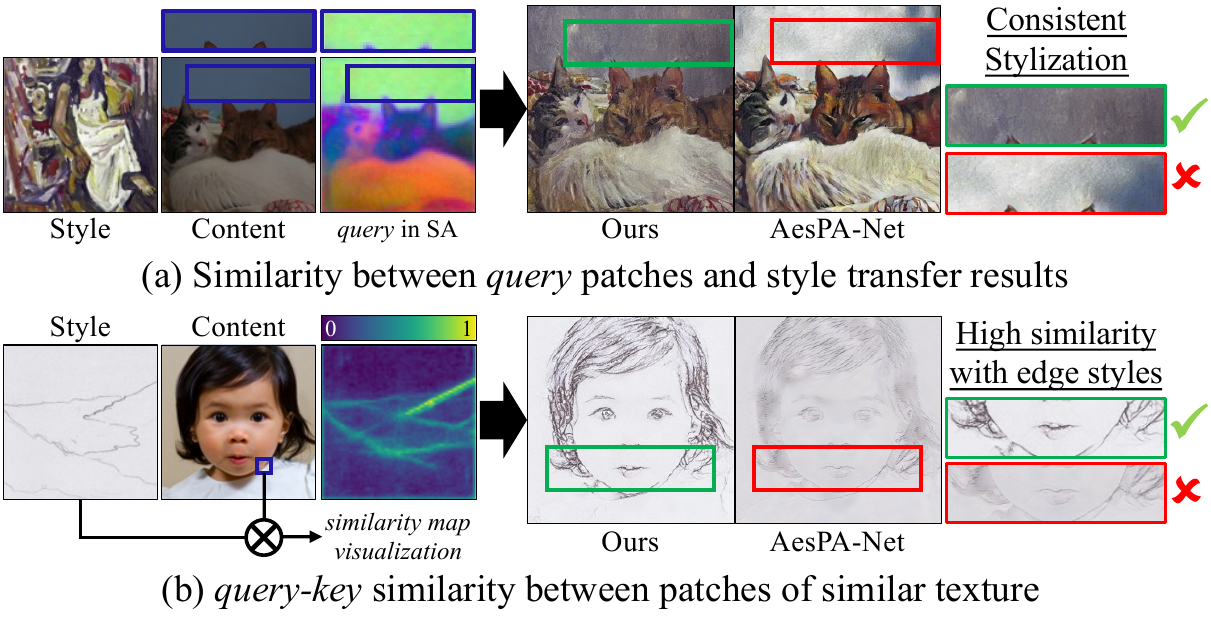}
    \centering
    \vspace*{-0.6cm}
    \caption{
    \textbf{Desirable attributes of self-attention~(SA) for style transfer.} (a) Visualization of \textit{query} by PCA shows that \textit{query} features well-reflect similarities among patches. 
    That is, style transfer employing SA can preserve the original content, as content patches with similarities tend to receive similar attention scores from a corresponding style image patch.
    (b) We visualize a similarity map between the blue box~(edge) \textit{query} of the content image, and \textit{key} of the style image.
    Thanks to the features representation of large-scale DM encompassing texture and semantics, a \textit{query} exhibits higher similarity to \textit{key}s that share a similar style, such as edges.
    }
    \vspace{-0.5cm}
    \label{fig_motivation_self_attn_attributes}
\end{figure}

In this paper, we focus on extending the training-free style transfer to its application on large-scale pre-trained DM.
We start from the observation of recent advances in image-to-image translation based on large-scale DM; they uncover the image editing capability of attention layers.
Notably, Plug-and-play~\cite{tumanyan2023plug} shows that the residual block and the attention map of self-attention~(SA) determine the spatial layout of generated images.
Also, Prompt-to-Prompt~\cite{hertz2022prompt} locally edits the image by replacing \textit{key} and \textit{value} of cross-attention~(CA) obtained from text prompt, while keeping their original attention maps.
That is, all these works suggest that 1) attention maps determine the spatial layout and 2) \textit{key} and \textit{value} of CA adjust the content to fill.

Inspired by the aforementioned methods, we newly argue that manipulating the SA layer is an effective way to transfer the styles~(Fig.~\ref{fig_motivation_self_attn_method}).
Specifically, similar to CA, we substitute the \textit{key} and \textit{value} of SA and observe that the generated images are still visually plausible and naturally incorporate the elements of the substituted image into the original image.
This observation motivates us to propose a style transfer technique based on self-attention, which combines the styles~(textures) of a specific image with the content~(semantics and spatial layout) of different images.
Furthermore, we highlight that SA layer has desirable characteristics for style transfer.
First, as shown in Fig.~\ref{fig_motivation_self_attn_attributes}~(a), in SA-based style transfer, the content image patches~(\textit{query}) that share semantic similarities engage with a similar style~(\textit{key}), thereby maintaining the relationship among these content image patches after the transfer.
Next, thanks to the powerful feature representation of large-scale DM~\cite{zhang2023tale}, each patch of the \textit{query} reveals higher similarity to \textit{key}s which has similar texture and semantics.
For instance, in Fig.~\ref{fig_motivation_self_attn_attributes}~(b), we can observe that the \textit{query} feature of content within the blue box exhibits a high similarity to the \textit{key} features of style with similar edge texture.
This encourages the model to transfer style based on the similarity of local texture~(e.g. edge) between content and style.

As a result, our method aims to transfer the textures of the style image to the content images by manipulating self-attention features of pre-trained large-scale DM without any optimization.
To this end, we first propose an attention-based style injection method.
The basic idea of it is substituting the content's \textit{key} and \textit{value} of SA with those of the style image, especially layers in the latter part of decoder which are relevant to the local textures.
As mentioned in above paragraph, exchanged styles are well aligned with the content and texture of original image, exploiting the similarity-based attention mechanism.
With the proposed style injection, we observe that the local texture patterns are successfully transferred, but there still are remaining problems such as disruption of original content and disharmonious colors.
To handle these problems, we additionally propose the following techniques; query preservation, attention temperature scaling, and initial latent AdaIN.
Query preservation makes the reverse diffusion process to retain the spatial structure of original content by preserving the \textit{query} of the content image in the SA.
Attention temperature scaling also aims to keep the structure of content by dealing with the blurred self-attention map introduced from the substitution of \textit{key}.
Lastly, initial latent AdaIN corrects inharmonious color problem, referring that the color distribution of style images is not properly transferred, by modulating the statistics of initial noise in the diffusion model.

Our main contributions are summarized as follows:
\begin{itemize}
    \item[--] 
    We propose a style transfer method exploiting the large-scale pre-trained DM by simple manipulation of the features in self-attention; substituting \textit{key} and \textit{value} of content with those of styles without any requirements of optimization or supervision~(e.g. text).
    \item[--] 
    We further improve the naive approach for style transfer to properly adapt the styles by proposing three components; query preservation, attention temperature scaling, and initial latent AdaIN.
    \item[--] 
    Extensive experiments on the style transfer dataset validate the proposed method significantly outperforms previous methods and achieves state-of-the-art performance.
\end{itemize}

\vspace{-0.1cm}
\section{Related Work}
\label{sec:related}
\subsection{Diffusion Model-based Neural Style Transfer}
Neural style transfer~\cite{gatys2016image, li2017universal, li2018closed, lai2017deep, park2019arbitrary, lu2019closed, wang2020collaborative, wang2020diversified, zhang2022domain} is an example-guided image generation task that transfers the style of one image onto another while retaining the content of the original.
In the realm of diffusion models, neural style transfer has evolved by leveraging the generative capability of pre-trained diffusion models.
For instance, InST~\cite{zhang2023inversion} introduced a textual inversion-based approach, aiming to map a given style into corresponding textual embeddings.
StyleDiffusion~\cite{wang2023stylediffusion} aimed to disentangle style and content by introducing CLIP-based style disentanglement loss for fine-tuning DM for style transfer.
Also, several approaches utilize the text input as a style condition or for determining the content to synthesize~\cite{everaert2023diffusion,yang2023zero}.

Conversely, DiffStyle~\cite{jeong2023training} proposed a training-free style transfer method that leverages \textit{h-space}~\cite{kwon2022diffusion} and adjusts skip connections for effectively conveying style and content information, respectively. 
However, when DiffStyle is applied to Stable Diffusion~\cite{rombach2022high, von-platen-etal-2022-diffusers}, their behavior is quite different from typical style-transfer methods; not only textures but also semantics such as spatial layout are also changed.

To address these limitations, we propose a novel algorithm that harmoniously merges style and content features within the self-attention layers of Stable Diffusion without any optimization process.

\subsection{Attention-based Image Editing in DM}
Following the remarkable advances achieved by pre-trained text-to-image DMs~\cite{ramesh2022hierarchical, von-platen-etal-2022-diffusers}, there have been numerous image editing works~\cite{avrahami2022blended, kawar2023imagic, couairon2023diffedit, shi2023dragdiffusion} utilizing these DMs.
Notably, Prompt-to-Prompt~\cite{hertz2022prompt} proposed text-based local image editing by manipulating the cross-attention map. Specifically, they observe that cross-attention largely contributes to modeling the relation between the spatial layout of the image to each word in the prompt. Hence, they substitute the original words and cross-attention map with desirable ones, obtaining edited images matched with text conditions. 
Subsequently, Plug-and-play~\cite{tumanyan2023plug} introduces text-guided image-to-image translation method. They found that the spatial features~(i.e. feature from residual block) and self-attention map determine the spatial layout of the synthesized image. Thus, while generating a new image with the given text condition, they guide the diffusion model with features and attention map from the original image for preserving the original spatial layout.
Recently, MasaCtrl~\cite{Cao_2023_ICCV} proposes mutual self-attention control for consistent image editing using text prompts. 
In detail, they retain the source image's $key$ and $value$ of the self-attention layers, while conditioning the model with desired text prompts.

Along with these works, we recognize the potential of attention maps in representing spatial information. 
However, different from the aforementioned methods concentrating on exploiting textual condition, we focus on conditioning by style and content images composed of two images from distinct styles.
By combining the features in self-attention layers of both style and content images with precise adjustment of statistics in intermediate representations, we transfer the texture of the content image to the given style.

\vspace{-0.1cm}
\section{Background}
\label{sec:background}
\begin{figure*}
    \centering
    \includegraphics[width=1.0\linewidth]{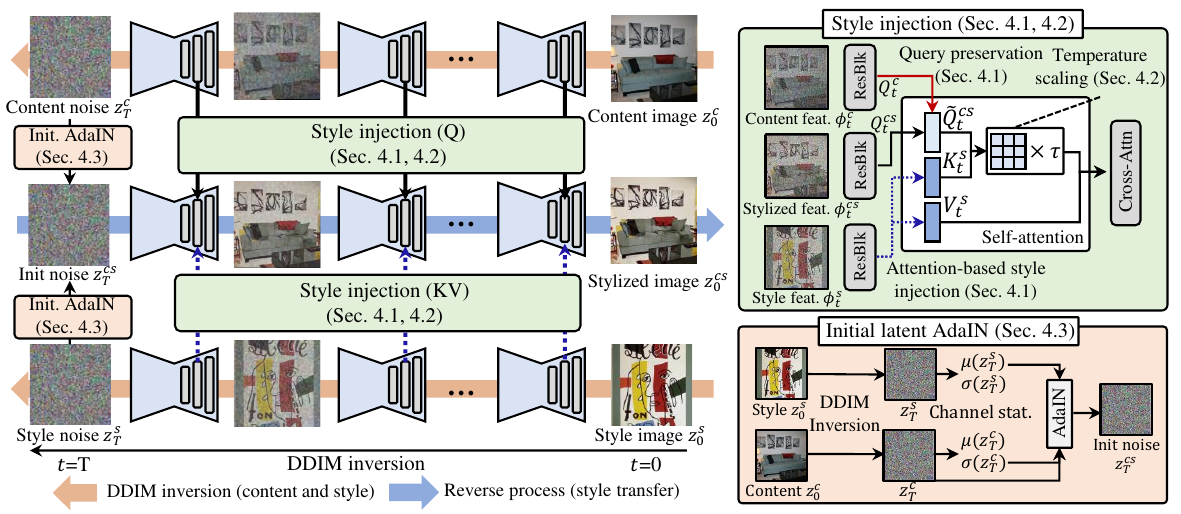}
    \vspace*{-0.6cm}
    \caption{\textbf{Overall framework.}~(Left) Illustration for the proposed style transfer method. We first invert content image $z^c_0$ and style image $z^s_0$ into the latent noise space as $z^c_T$ and $ z^s_T$, respectively. 
    Then, we initialize the initial noise of stylized image $z_T^{cs}$ from initial latent AdaIN~(Sec.~\ref{sec_init_adain}) which combines the content and style noise, $z_T^c$ and $z_T^s$.
    While performing the reverse diffusion process with $z^{cs}_T$, we inject the information of content and style by attention-based style injection~(Sec.~\ref{sec_attn_fusion}) and attention temperature scaling~(Sec.~\ref{sec_attn_rev_tem}).
    (Right) Detailed explanation of style injection and initial noise AdaIN. Style injection is basically the manipulation of self-attention~(SA) layer during the reverse diffusion process. Specifically, at time step $t$, we substitute the \textit{key}~($K^{cs}_t$) and \textit{value}~($V^{cs}_t$) in SA of stylized image with those of style features, $K^s_t$ and $V^s_t$, from identical timestep $t$. At the same time, we preserve the content information by blending the \textit{query} of content $Q^c_t$ and \textit{query} of stylized image $Q^{cs}_t$. Finally, we scale the magnitude of the attention map to deal with the magnitude decrease that the substitution of feature leads to. 
    Initial latent AdaIN produces the initial noise $z_T^{cs}$ by combining style noise $z_T^s$ and content noise $z_T^s$.
    Specifically, we modify the channel statistics of $z_T^c$ to resemble the statistics of $z_T^s$ and regard it as $z_T^{cs}$.
    We observe this operation enables us to keep the spatial layout of content image while well-reflecting the color tones of a given style image.
    }
    \vspace*{-0.4cm}
    \label{fig_overall_framework}
\end{figure*}
Latent Diffusion Model~(LDM)~\cite{rombach2022high} is a type of diffusion model trained in the low dimensional latent space to focus on semantic bits of data and reduce computation costs.
Given an image $x\in\mathbb{R}^{H\times W\times 3}$, the encoder~$\mathcal{E}$ encodes $x$ into the latent representation $z\in\mathbb{R}^{h\times w\times c}$ and decoder reconstructs the image from the latent.

With the pretrained encoder, they encode the entire images in the dataset and train a diffusion model on latent space $z$, by predicting noise $\epsilon$ from the noised version of latent $z_t$ at time step $t$.
The corresponding training objective is 
\begin{equation}
    \label{ldm_equation}
    L_\text{LDM}=\mathbb{E}_{z,\epsilon,t}[\Vert\epsilon-\epsilon_\theta(z_t, t, y)\Vert^2_2],
\end{equation}
where $\epsilon\in\mathcal{N}(0, 1)$ is a noise, $t$ is the number of time steps which uniformly sampled from $\{1,..., T\}$, $y$ is a condition, and $\epsilon_\theta$ is a neural network which predicts the noise added to the $z$.

In our work, we utilize Stable Diffusion~(SD)~\cite{rombach2022high} which is the only publicized large-scale pre-trained DM. 
In the case of SD, $y$ is a text, and $\epsilon_\theta$ is a U-Net architecture in which a block for each resolution comprises a residual block, self-attention block~(SA), and cross-attention block~(CA), sequentially.
Among these modules, we focus on the SA block to transfer the styles, as discussed in Sec.~\ref{sec:intro}.
Given a feature $\phi$ after the residual block, the self-attention block performs as follows:
\begin{equation}
    \label{self_attn_equation}
    \begin{split}
    Q = W_Q(\phi), K&=W_K(\phi), V=W_V(\phi), \\
    \phi_\text{out} = \text{Attn}(Q, K, V) & = \text{softmax}(\frac{Q {K}^{T}}{\sqrt{d}})\cdot V,
    \end{split}
\end{equation}
where $d$ denotes the dimension of the projected \textit{query}, and $W_{(\cdot)}$ is a projection layer.
Note that, we don't use any text conditions, so the variable $y$ is always an empty text prompt~(``").

\vspace{-0.1cm}

\section{Method}
\label{sec:method}

In this paper, we aim to solve artistic style transfer by leveraging the generative capability of a pre-trained large-scale text-to-image diffusion model.
Briefly, artistic style transfer is the task of modifying the style of a given content image $I^c$ to that of style image $I^s$.
Then, the stylized image $I^{cs}$ would maintain the semantic content of $I^c$ while its style~(such as texture) is transferred from $I^s$. 
For simplicity, we skip the explanations about the encoding and decoding process of the autoencoder in the LDM. Instead, we focus on elaborating the proposed method in the aspect of the diffusion process.
Thus, in the following sections, we regard the content, style, and stylized images same as their encoded counterparts $z^c_0$, $z^s_0$, and $z^{cs}_0$.

\subsection{Attention-based Style Injection}
\label{sec_attn_fusion}
We start from the observation in previous image-to-image translation methods, especially Prompt-to-Prompt~\cite{hertz2022prompt}.
The key idea of their method is changing the text condition for cross-attention~(CA) while keeping the attention map.
Since the attention map affects the spatial layout of output, substituted text conditions determine what to draw in the generated image, and these conditions are actually \textit{key} and \textit{value} in CA.
Inspired by them, we manipulate the features in self-attention layer as like cross-attention, regarding the features from style image $I^s$ as the condition.
Specifically, in the generation process, we substitute the \textit{key} and \textit{value} of content image with those of style for transferring the texture of style image into the content image.

To this end, we first obtain the latent for content and style images with DDIM inversion~\cite{song2020denoising}, and then collect the SA features of style image over the DDIM inversion process.
Specifically, for pre-defined timesteps $t=\{0, ..., T\}$, style and content images $z^c_0$ and $z^s_0$ are inverted from image~($t=0$) to gaussian noise~($t=T$).
During DDIM inversion, we also collect \textit{query} features of content~($Q^c_t$) and \textit{key} and \textit{value} features of style~($K^s_t, V^s_t$) at every time steps.

After that, we initialize stylized latent noise $z^{cs}_T$ by copying content latent noise $z^c_T$.
Then, we transfer the target style to the stylized latent by injecting the \textit{key} $K^s_t$ and \textit{value} $V^s_t$ collected from the style into SA layer, instead of the original \textit{key} $K^{cs}_t$ and \textit{value} $V^{cs}_t$, when performing the entire reverse process of stylized latent $z^{cs}_t$.
However, only applying this substitution can lead to content disruption, since the content of stylized latent would be progressively changed as attended \textit{value} changes.
Hence, we propose query preservation to maintain original content.
Simply, we blend \textit{query} of stylized latent $Q^{cs}_t$ and that of content $Q^{c}_t$ for the entire reverse process.
These style injection and query preservation processes at time step $t$ are expressed as follows:
\begin{equation}
    \label{eq_attn_query_mix}
    \Tilde{Q}_t^{cs}=\gamma \times Q_t^c + (1-\gamma) \times Q_t^{cs},
\end{equation}
\begin{equation}
    \label{eq_attn_injection}
    \phi^\text{cs}_\text{out} = \text{Attn}(\Tilde{Q}_t^{cs},K_t^s,V_t^s),
\end{equation}
where $\gamma$ is degree of blending in range of $[0, 1]$.
In addition, we apply these operations on the latter layers of decoder~(7-12$^{\text{th}}$ decoder layers in SD) relevant to local textures.
We also highlight that the proposed method can adjust the degree of style transfer by changing query preservation ratio $\gamma$.
Specifically, higher $\gamma$ maintains more content, while lower $\gamma$ strengthens effects of style transfer.

\subsection{Attention Temperature Scaling}
Attention map is computed by scaled dot-product between \textit{query} and \textit{key} features.
During training, \textit{query} and \textit{key} features in the SA layer originate from an identical image.
However, if we substitute the key features with those of style images, the magnitude of similarity would be overall lowered as style and content are highly likely to be irrelevant.
Thus, the computed attention map can be blurred or smoothed, and it would further make output images unsharp, which is detrimental to capturing both content and style information.

\begin{figure}[t!]
     \begin{subfigure}[b]{0.265\textwidth}
     \centering
     \includegraphics[width=\textwidth]{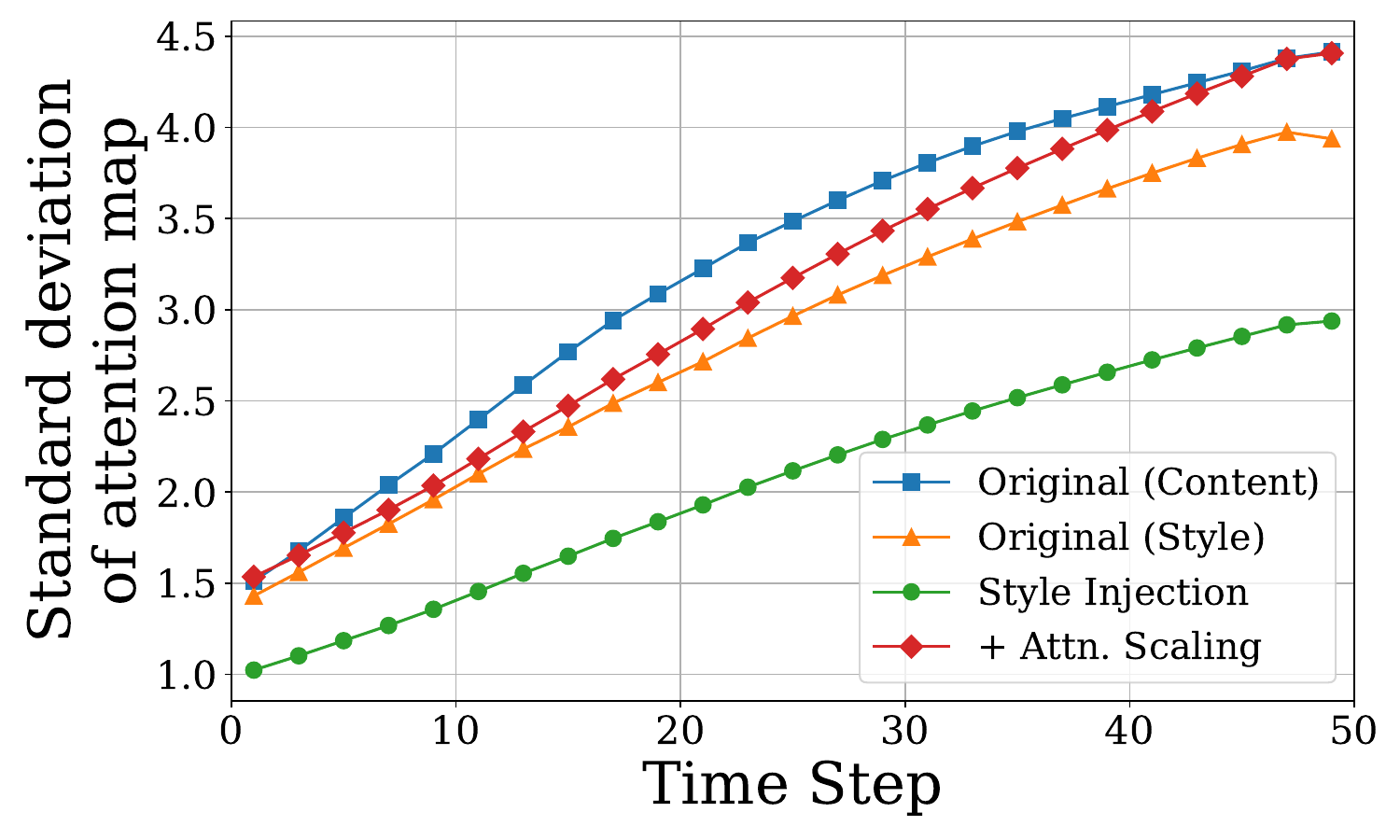}
     \caption{Std. of attention map}
     \label{fig_attnmap_std}
 \end{subfigure}
 \hspace*{-0.25cm}
 \begin{subfigure}[b]{0.215\textwidth}
     \centering
     \includegraphics[width=\textwidth]{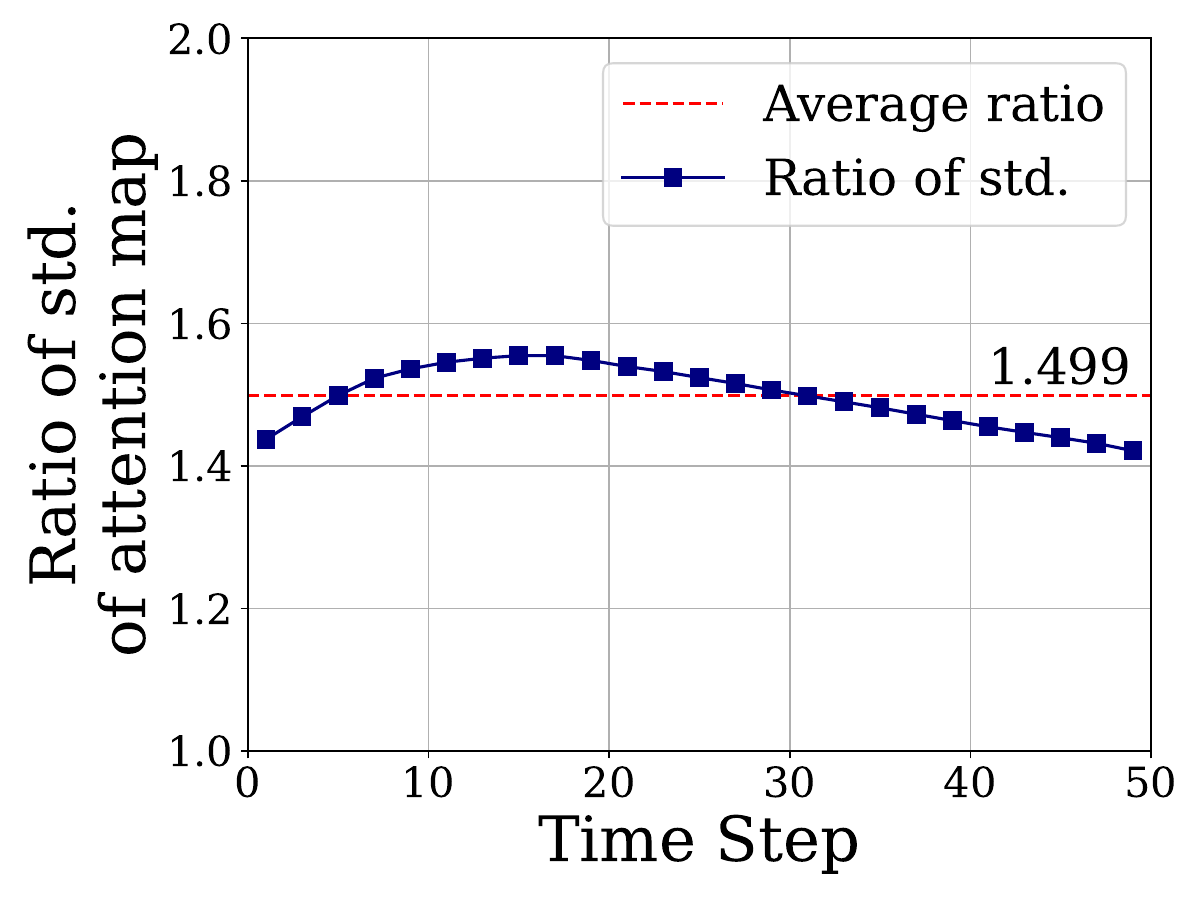}  
     \caption{Ratio of std. of attention map}
     \label{fig_attnmap_std_ratio}
 \end{subfigure}
    \centering
    \vspace{-0.6cm}
    \caption{
    \textbf{Visualization of the standard deviation of attention map before softmax.}
    (a) Attention-based style injection reduces the standard deviation of self-attention map. 
    Original denotes SA maps from the generation process without style injection. We use both style and content images for generation.
    (b) We compute the ratio between attention maps w/ and w/o style injection. For the std of original image, we use averaged std. of content and style.
    }
    \vspace{-0.4cm}
    \label{fig_bluriness}
\end{figure}

To quantify this issue, we measure the standard deviation of attention map, while ablating the attention-based style injection.
In detail, we calculate the attention map before applying softmax, which is scaled-dot product between \textit{query} and \textit{key}.
As shown in Fig.~\ref{fig_bluriness}~(a), we validate that this style injection tends to lower the standard deviation of the attention map over the entire timesteps.
That is, attention maps after softmax with style injection would be overly smooth.

To rectify the attention map sharper, we introduce an attention temperature scaling parameter.  
In detail, we multiply the attention map before softmax by a constant temperature scaling parameter $\tau$ larger than 1. Thus, the attention map after softmax would be sharper than its original values.
The modified attention process is represented as follows:
\begin{equation}
    \label{eq_attn_rev_tem}
    \text{Attn}_{\tau}(\Tilde{Q^{cs}_t},K^s_t,V^s_t)=\text{softmax}(\frac{\tau~ \Tilde{Q^{cs}_t} ({K^s_t})^{T} }{ \sqrt{d}})\cdot V^s_t, \tau>1.
\end{equation}

We use $\tau=1.5$ as a default setting, which is the average ratio over entire timesteps. As reported in Fig.~\ref{fig_bluriness}~(b), we confirm that it effectively calibrates the standard deviation of attention map similar to its original values.

\label{sec_attn_rev_tem}

\begin{figure}[t!]
    \includegraphics[width=0.9\columnwidth]{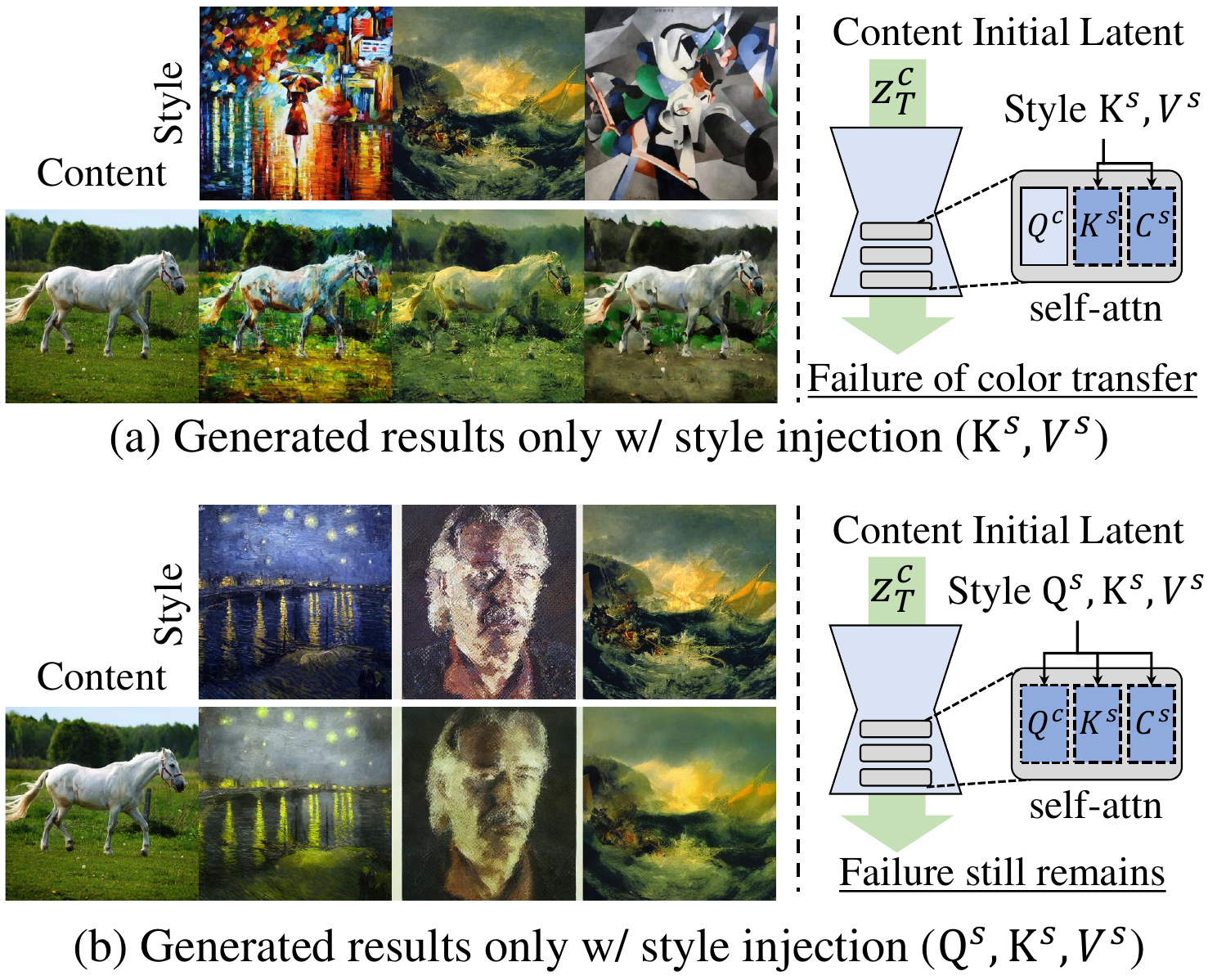}
    \centering
    \vspace{-0.2cm}
    \caption{
    \textbf{Generated results only w/ style injection.}
    (a) We observe that generated images only with attention-based style injection do not harmonize with the given style in the aspect of color tone. 
    (b) To identify the effects of every feature in SA on color tones, we additionally include \textit{query} in the style injection process. However, color tones still resemble those of content, concluding features in self-attention have less effect on the color tones. 
    }
    \vspace{-0.4cm}
    \label{fig_color_tone}
\end{figure}

\begin{table*}
\centering
\footnotesize
\setlength{\tabcolsep}{4pt} % 간격을 조절하려면 이 부분을 조절
\begin{tabular}{c|cccccccccc|ccc}
\toprule
% \renewcommand{\arraystretch}{0.8} % row 간격
% aespa-net iccv 2023
% CAST SIGGRAPH 2022
% stytr2 cvpr 2022
% EFDM cvpr 2022
% MAST iccv 2021
% AdaAttn iccv 2021
% artflow cvpr 2021
% AdaConv cvpr 2021
% AdaIN iccv 2017???
\textbf{Metric} & \textbf{Ours} & \textbf{AesPA-Net} & \textbf{CAST} & \textbf{StyTR$^2$} & \textbf{EFDM} & \textbf{MAST} & \textbf{AdaAttn} & \textbf{ArtFlow} & \textbf{AdaConv} & \textbf{AdaIN} & \textbf{DiffuseIT} & \textbf{InST} & \textbf{DiffStyle} \\
\midrule
ArtFID $\downarrow$ & \textbf{28.801}   & 31.420 & 34.685 & 30.720 & 34.605 & 31.282 & 30.350 & 34.630 & 31.856 & 30.933 & 40.721 & 40.633 & 41.464 \\
FID $\downarrow$    & \textbf{18.131}   & 19.760 & 20.395 & 18.890 & 20.062 & 18.199 & 18.658 & 21.252 & 19.022 & 18.242 & 23.065 & 21.571 & 20.903 \\
LPIPS $\downarrow$  & \textbf{0.5055}   & 0.5135 & 0.6212 & 0.5445 & 0.6430 & 0.6293 & 0.5439 & 0.5562 & 0.5910 & 0.6076 & 0.6921 & 0.8002 & 0.8931 \\
\midrule
CFSD $\downarrow$   & \textbf{0.2281}   & 0.2464 & 0.2918 & 0.3011 & 0.3346 & 0.3043 & 0.2862 & 0.2920 & 0.3600 & 0.3155 & 0.3428 & 0.6759 & 0.2819 \\
\bottomrule

\end{tabular}
    \vspace{-0.2cm}
    \caption{
        Quantitative comparison with conventional~(3$^{\text{rd}}$-11$^{\text{th}}$ columns) and diffusion model baselines~($12^{\text{th}}$-$14^{\text{th}}$ columns)
    }
    \vspace{-0.4cm}
\label{tab_quan_con}
\end{table*}

\subsection{Initial Latent AdaIN}
\label{sec_init_adain}
\Skip{
    In conventional artistic style transfer~\cite{}, the integration of color tone into the style information is a common practice. However, even after implementing attention-based style injection, the output image often fails to capture the color tone of the style image, as shown in Fig.~\ref{}.
}
In artistic style transfer, the color tone generally takes up a significant portion of the style information.
In this context, we observe that the style transfer only with attention-based style injection often fails in terms of capturing the color tone of the given style.
As shown in Fig.~\ref{fig_color_tone}~(a), textures and local patterns are successfully transferred to the content image while the color tone of the content image still remains.
Furthermore, even with injecting the \textit{query}, \textit{key}, and \textit{value} of styles, the resulting images still preserve the color tone of the content,  as shown in Fig.~\ref{fig_color_tone}~(b).

As substituting the self-attention features has less effects to color tone, we analyze the other vital part of DM; initial latent noise.
One of the recent discoveries in DM is that the DM struggles to synthesize purely white or black images~\cite{nicholas2023offsetnoise}.
Instead, they tend to generate images of median color as the initial noise is sampled from zero mean and unit variance.
Thus, we hypothesize the statistics of initial noise largely affect the colors and brightness of generated images.

Based on this assumption, we attempt to use the initial latent of style $z^s_T$ for the style transfer process.
However, if we simply start to generate the image from style latent $z^s_T$, the structural information of synthesized results also follows the style image and loses the structure of the content.
To harness valuable information in both initial latents, we consider that the tone information is intricately connected with the channel statistics of the initial latent, following the principle underlying Style Loss~\cite{gatys2016image} and AdaIN~\cite{huang2017arbitrary}. 
Thus, we employ AdaIN to modulate the initial latent for effective tone information transfer, represented as:
\begin{equation}
\label{eq_init_adain}
z^{cs}_T=\sigma(z^s_T)\left(\frac{z^c_T-\mu(z^c_T)}{\sigma(z^c_T)} \right) + \mu(z^s_T),
\end{equation}
where $\mu(\cdot),\sigma(\cdot)$ denote channel-wise mean and standard deviation, respectively.
Based on this, the initial latent $z^{cs}_T$ preserves content information from $z^c_T$ while aligning the channel-wise mean and standard deviation with $z^s_T$.

\section{Experiments}
\label{sec:experiments}

\subsection{Experimental Settings}
We conduct all experiments in Stable Diffusion~1.4 pre-trained on LAION dataset~\cite{schuhmann2022laion} and adopt DDIM sampling~\cite{song2020denoising} with a total 50 timesteps~($t=\{1, ..., 50\}$).  
For default settings for hyperparameters, we use $\gamma=0.75$ and $\tau=1.5$, if they are not mentioned separately.

\subsection{Evaluation Protocol}
Conventional style transfer methods typically utilize Style Loss~\cite{gatys2016image} as both training objective and evaluation metric, so their results tend to overfit the Style Loss.
Thus, for a fair comparison, we employ a recently proposed metric, ArtFID~\cite{wright2022artfid} which evaluates overall style transfer performances with consideration of both content and style preservation and also is known as strongly coinciding with human judgment.
Specifically, ArtFID is computed as $(\text{ArtFID}=(1+\text{LPIPS})\cdot(1+\text{FID}))$. 
LPIPS~\cite{zhang2018unreasonable} measures content fidelity between the stylized image and the corresponding content image, and FID~\cite{heusel2017gans} assesses the style fidelity between the stylized image and the corresponding style image.
% strongly coincides with human judgment.

\vspace{0.2cm}
% \paragraph{Dataset}
\noindent\textbf{Dataset.}
Our evaluations employ content images from \textit{MS-COCO}~\cite{lin2014microsoft} dataset and style images from \textit{WikiArt}~\cite{artgan2018} dataset. All input images are center-cropped to 512 $\times$ 512 resolution.
Also, for quantitative comparison, we randomly selected 20 content and 40 style images from each dataset, yielding 800 stylized images as StyTR$^2$~\cite{deng2022stytr2} has done.

\vspace{0.2cm}
\noindent\textbf{Content Feature Structural Distance~(CFSD). }
In the style transfer evaluation, the assessment of content fidelity often relies on the LPIPS distance.
However, since LPIPS utilizes the feature space of AlexNet~\cite{krizhevsky2012imagenet} pre-trained for classification task on ImageNet~\cite{deng2009imagenet}, which is known as texture-biased~\cite{geirhos2018imagenet}.
Thus, the style information of the images can affect the LPIPS score.
To mitigate this style influence, we additionally introduce Content Feature Structural Distance~(CFSD) which is a distance measure that only considers the spatial correlation between image patches.

In detail, we first define the correlation map between image patch features as follows.
For a given image $I$, we obtain feature maps $F\in\mathbb{R}^{hw\times c}$, which is the output feature of \textit{conv3} in VGG19~\cite{simonyan2014very}. 
Then, we calculate the patch similarity map $M=F \times F^{T},M\in\mathbb{R}^{hw \times hw}$, which is a similarity map between every pair of features in $F$.
After that, for computing the distance between two patch similarity maps, we model the similarity between a single patch and the others as a probability distribution by applying softmax operation.
Finally, the correlation map is represented as $S=[\text{softmax}(M_i)]^{hw}_{i=1}, S\in\mathbb{R}^{hw \times hw}$, where $M_i\in\mathbb{R}^{1 \times hw}$ is a similarity map between $i^{\text{th}}$ patch and the other patches.
% i번째 patch가 similarity map of i^th patch
% \SE{} % M_i 정의 필요
% \JW{} % 수정 완료 

Then, CFSD is defined as KL-divergence between two correlation maps.
In our case, we compute CFSD between the correlation map of the content~($S^c$) and stylized images~($S^{cs}$) as follows:
\begin{equation}
\label{eq_s2sim}
\text{CFSD} = \frac{1}{hw} \sum^{hw}_{i=1}D_\text{KL}(S^c_i||S^{cs}_i),
\end{equation}

\begin{figure*}[t!]
    \vspace{-0.1cm}
    \includegraphics[width=1.0\textwidth]{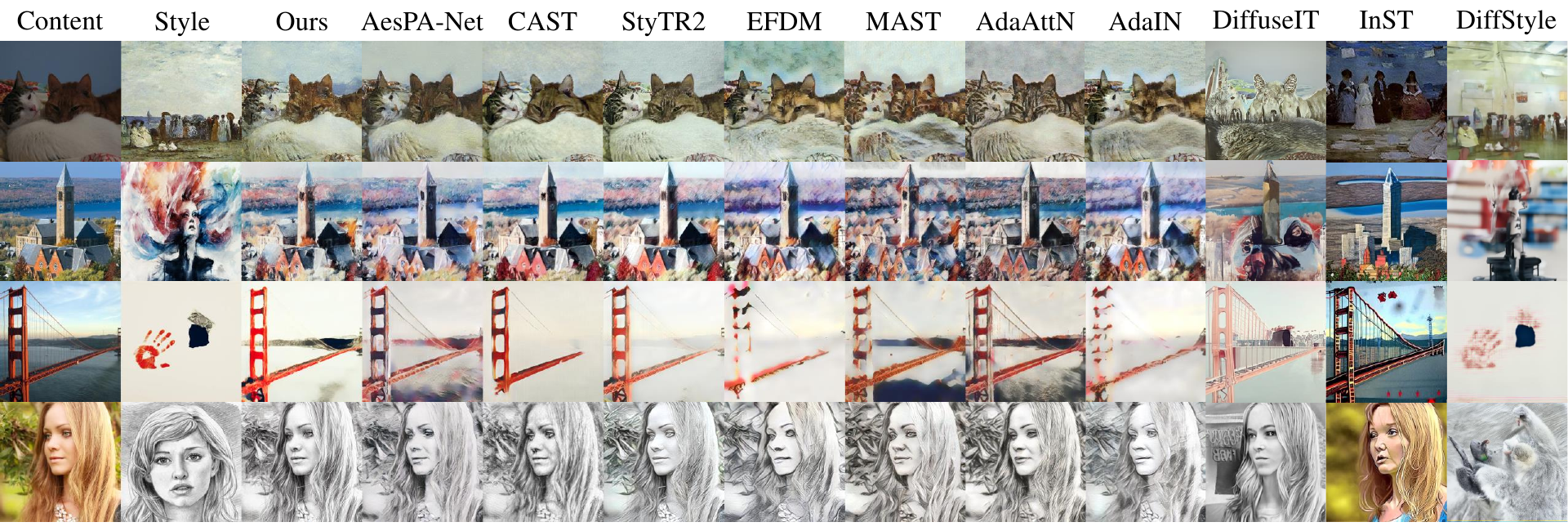}
    \centering
    \vspace*{-0.8cm}
    \caption{
        Qualitative comparison with conventional~(4$^{\text{th}}$-10$^{\text{th}}$ columns) and diffusion model baselines~($11^{\text{th}}$-$13^{\text{th}}$ columns)
    }
    \vspace{-0.5cm}
    \label{fig_qualitative}
\end{figure*}

\begin{figure}[t!]
    \vspace{-0.1cm}
    \includegraphics[width=0.95\columnwidth]{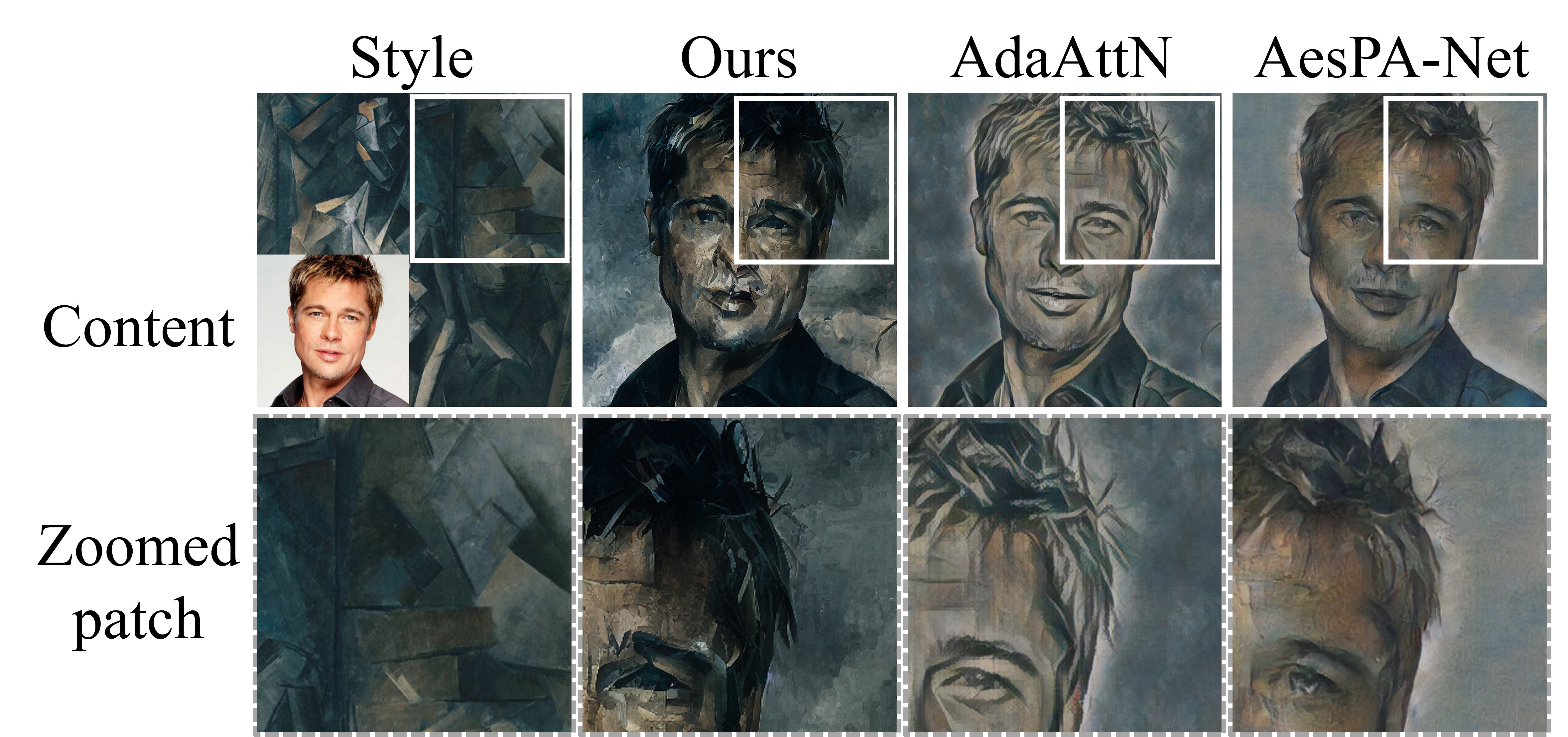}
    \centering
    \vspace{-0.3cm}
    \caption{
        Qualitative comparison with best ArtFID~(AdaAttN) and most recently proposed baselines~(AesPA-Net) with additional zooming details
    }
    \vspace{-0.2cm}
    \label{fig_qualitative_cropped}
\end{figure}

\subsection{Quantitative Comparison}
We evaluate our proposed method through comparison with twelve state-of-the-art methods, including nine conventional style transfer methods~(AesPA-Net~\cite{hong2023aespa}, CAST~\cite{zhang2022domain}, StyTR$^2$~\cite{deng2022stytr2}, EFDM~\cite{zhang2022exact}, MAST~\cite{deng2020arbitrary}, AdaAttN~\cite{liu2021adaattn}, ArtFlow~\cite{an2021artflow}, AdaConv~\cite{chandran2021adaptive}, AdaIN~\cite{huang2017arbitrary}) and three diffusion-based style transfer methods~(DiffuseIT~\cite{kwon2022diffuseit}, InST~\cite{zhang2023inversion}, DiffStyle~\cite{jeong2023training}), which have a style image as input.
We employ the publicly available implementations of all baselines, using their recommended configurations.

\vspace{0.2cm}
\noindent\textbf{Comparison with Conventional Style Transfer.}
As shown in Tab.~\ref{tab_quan_con}, our method largely surpasses the conventional style transfer methods in terms of ArtFID, which is known as coinciding the human preference.
In addition, the proposed method records the lowest FID, which denotes that stylized images highly resemble the target styles.
For content fidelity metrics, ours shows superior scores in both CFSD and LPIPS.
We point out that ours achieves much lower CFSD compared to other methods, which is the metric to only consider the spatial correlation. 

In addition, we also emphasize that the proposed method can arbitrarily adjust the degree of style transfer by changing the $\gamma$, and the proposed method significantly surpasses all the other methods in terms of FID~(style), when we match the value of LPIPS~(content)~(Fig.~\ref{fig_trade_off_cs}).

\vspace{0.2cm}
\noindent\textbf{Comparison with Diffusion-based Style Transfer.}
Our method demonstrates the best performance in terms of LPIPS, FID, and their combination~(ArtFID) with a large margin, as shown in Tab.~\ref{tab_quan_con}.
One significant factor for diffusion models is their running times, since they require several steps to synthesize a single image and it requires inevitable time cost.
Hence, we measure the inference time for a pair of content and style images on a single TITAN RTX GPU, as shown in Tab.~\ref{tab_quan_inference_time}.
Our method requires a total of 12.4 seconds, with 8.2 seconds for DDIM inversions and 4.2 seconds for sampling costs.
As reported, we validate the proposed method significantly faster than other methods, even exploiting large-scale DM.
This faster speed comes from the fact that the proposed methods can use the much smaller steps of DDIM inversion, because we additionally utilize the features collected during inversion steps, largely reducing the necessity for perfect inversion of content and style.

\begin{table}
\centering
\footnotesize
\setlength{\tabcolsep}{4pt} % 간격을 조절하려면 이 부분을 조절
\begin{tabular}{c|cccc}
\toprule
Metric & DiffuseIT & InST & DiffStyle & Ours \\ 
\midrule
Time~(sec) & 792.8 & 816.9 & 355.9 & \textbf{12.4} \\
\bottomrule
\end{tabular}
\vspace{-0.2cm}
% \caption{Quantitative comparison with diffusion-based baselines}
\caption{Comparison of inference time of diffusion-based methods for style transferring a given style and content pair}
\vspace{-0.5cm}
\label{tab_quan_inference_time}
\end{table}

% 792.8336288928986
% 166.1 + 167.2 + 22.6

\subsection{Qualitative Comparison}
\noindent\textbf{Comparison with Conventional Style Transfer.}
As shown in Fig.~\ref{fig_qualitative}, we observe that our method tends to highly preserve the structural information of the content image, while also transferring the style well.
For instance, as shown in the third row, ours retains the structure of the bridge, but the baselines struggle to preserve structure or transfer the style.
We also provide the qualitative comparison with zooming details in Fig.~\ref{fig_qualitative_cropped} and Supplementary.

\vspace{0.2cm}
\noindent\textbf{Comparison with Diffusion-based Style Transfer.}
We also compare our method with recent diffusion-based style transfer baselines~\cite{zhang2023inversion, jeong2023training, kwon2022diffusion}. 
As shown in Fig.~\ref{fig_qualitative}, we observe the proposed technique transfers the style to the content well. 
On the other hand, baselines often lose the structure of content or fail to transfer the style, when an arbitrary content style pair is given.
For instance, DiffuseIT and DiffStyle suffer from generating shape and visually plausible images or drop the original content.
Differently, InST synthesizes the realistic images, while struggling to transfer style~($1^{\text{st}}$ row) or change content of image~($2^{\text{nd}}, 3^{\text{rd}}$ rows).

\subsection{Ablation Study}
\begin{figure}[t!]
    \includegraphics[width=1.0\columnwidth]{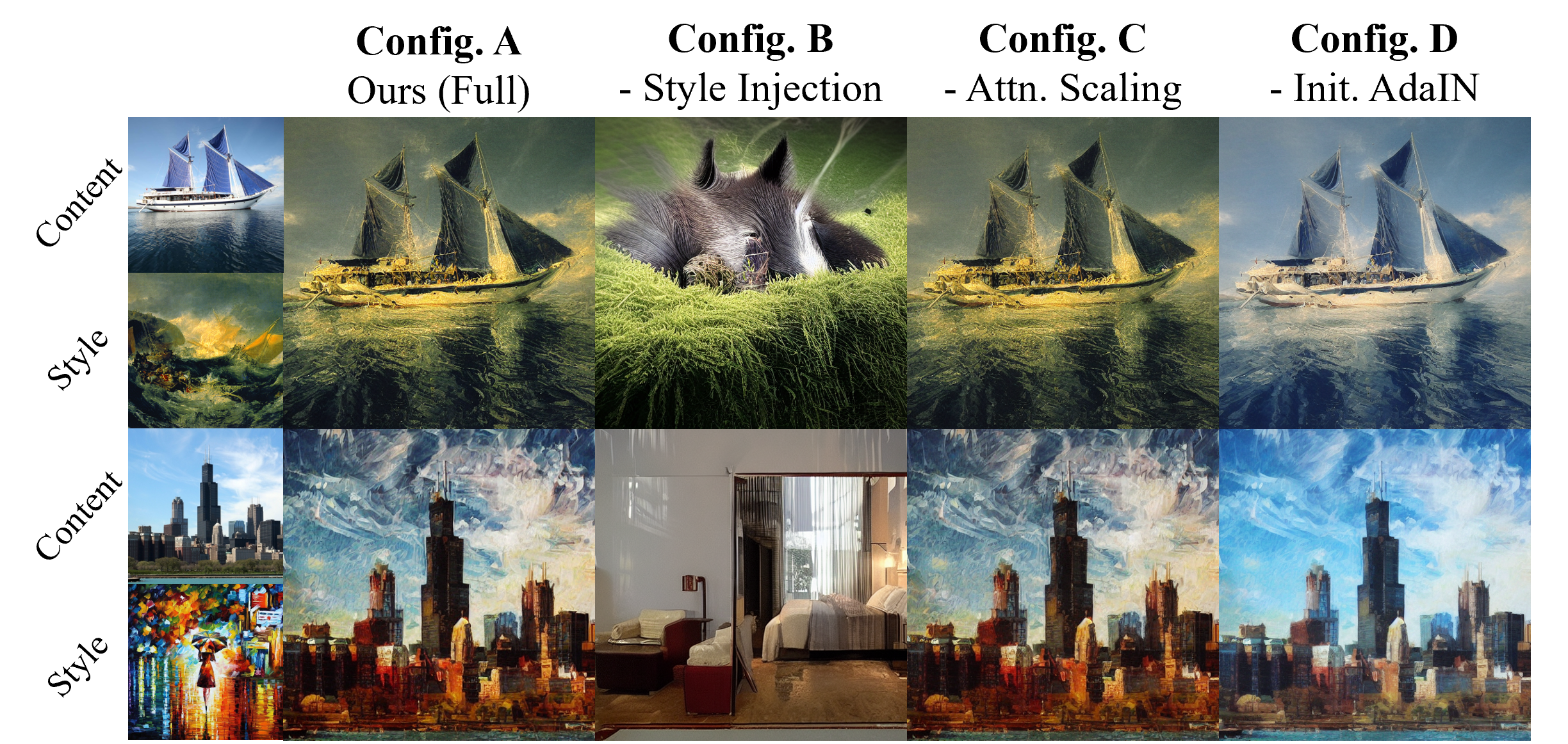}
    \centering
    \vspace{-0.6cm}
    \caption{
    Qualitative comparison with ablation studies
    }
    \vspace{-0.5cm}
    \label{fig_ablation}
\end{figure}

To validate the effectiveness of the proposed components, we conduct ablation studies in both quantitative and qualitative ways.
As shown in Fig.~\ref{fig_ablation} and Tab.~\ref{tab_ablation_study}, style injection is significant for guiding the style and content of given images~(Config.~B).
Besides, initial latent AdaIN has a large portion of transferring the color tone of style~(Config.~D).
Attention temperature scaling is in charge of enhancement of quality in synthesized results such as sharpening details and resolving blurriness.
For instance, this scaling jointly reduces the FID and LPIPS~(Config.~A$^*$~vs.~C in Tab.~\ref{tab_ablation_study}).
For more detailed analysis, we provide quantitative metrics with the style-content trade-off, while changing the attention scaling parameter $\tau$ in Fig.~\ref{fig_trade_off_cs}~(b).
As reported, attention scaling effectively reduces both FID and LPIPS, proving its effects on the preservation of content and capability of style transfer~($\tau=1.0$ vs. $\tau=1.5$).

\begin{table}[t!]
\centering
\begin{tabular}{cl|ccc}
\toprule
& \textbf{Configuration} & ArtFID & FID & LPIPS \\
\bottomrule
\arch{a} & Ours~\footnotesize{($\gamma=0.75$, default)} & 28.80 & 18.13 & 0.505 \\
\arch{a$^*$} & Ours~\footnotesize{($\gamma=0.6$)} & \textbf{27.97} & \textbf{17.21} & 0.535 \\
\bottomrule
\arch{b} & -~Style Inject.~\footnotesize{(Sec.~\ref{sec_attn_fusion})}   & 43.72 & 27.13 & 0.554 \\
\arch{c} & -~Attn. Scaling~\footnotesize{(Sec.~\ref{sec_attn_rev_tem})}    & 29.02 & 17.81 & 0.542 \\
\arch{d} & -~Init. AdaIN~\footnotesize{(Sec.~\ref{sec_init_adain})}        & 29.26 & 20.05 & \textbf{0.390} \\
\bottomrule
\end{tabular}
\vspace{-0.2cm}
\caption{Ablation study on proposed components}
\vspace{-0.4cm}
\label{tab_ablation_study}
\end{table}

\subsection{Additional Analysis}
\begin{figure}[t!]
    \centering
     \hspace*{-0.4cm}
     \begin{subfigure}[b]{0.31\textwidth}
         \centering
         \includegraphics[width=\textwidth]{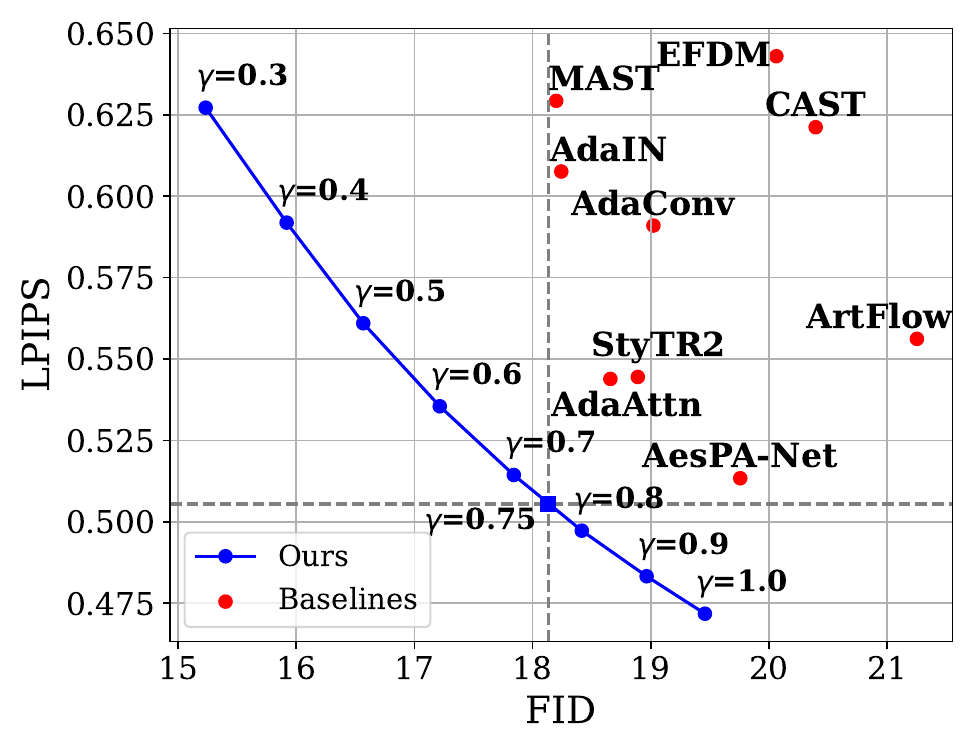}
         \vspace*{-0.65cm}
         \caption{Comparison with baselines}
         \label{fig_trade_off_baseline}
     \end{subfigure}
     \hspace*{-0.3cm}
     \begin{subfigure}[b]{0.19\textwidth}
         \centering
         \includegraphics[width=\textwidth]{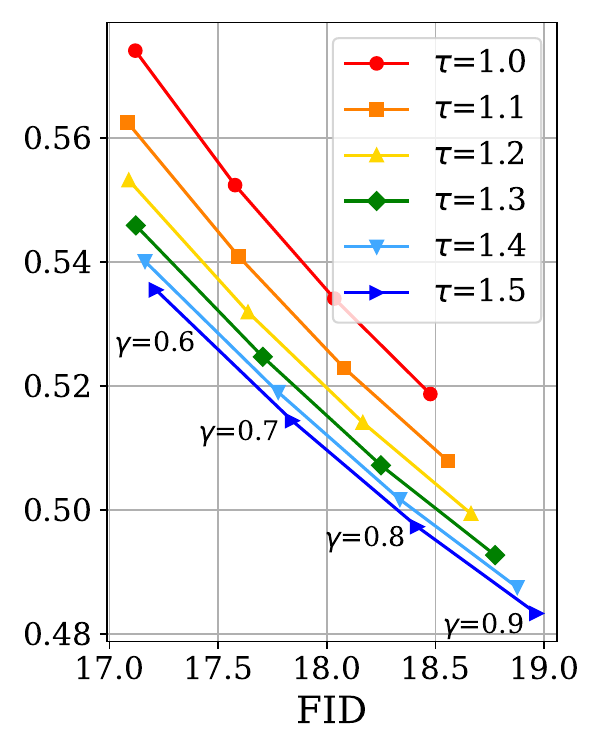}  
         \vspace*{-0.65cm}
         \caption{Comparison between $\tau$}
         \label{fig_trade_off_scaling}
     \end{subfigure}
     \vspace*{-0.7cm}
     \caption{Style-content trade-offs}
     \vspace{-0.2cm}
    \label{fig_trade_off_cs}
\end{figure}

\noindent\textbf{Content-Style Trade-Off.}
Our proposed method offers flexible control of the trade-off relation between content and style fidelity by adjusting the parameter $\gamma$, as discussed in Sec.~\ref{sec_attn_fusion}. 
In detail, we compute the FID and LPIPS while varying $\gamma$ within the range~$[0.3,1]$ with a step size of 0.1.
As shown in Fig.~\ref{fig_trade_off_cs}~(a), our method surpasses baseline methods across all ranges of content and style fidelity.
This result implies the proposed method significantly outperforms the other methods, when we match style or content metric to the compared model by adjusting the $\gamma$ of ours.
Note that, dotted lines refer to our model reported in Tab.~\ref{tab_quan_con}.

\begin{figure}[t!]
    \hspace*{-0.4cm}
    \includegraphics[width=1.0\columnwidth]{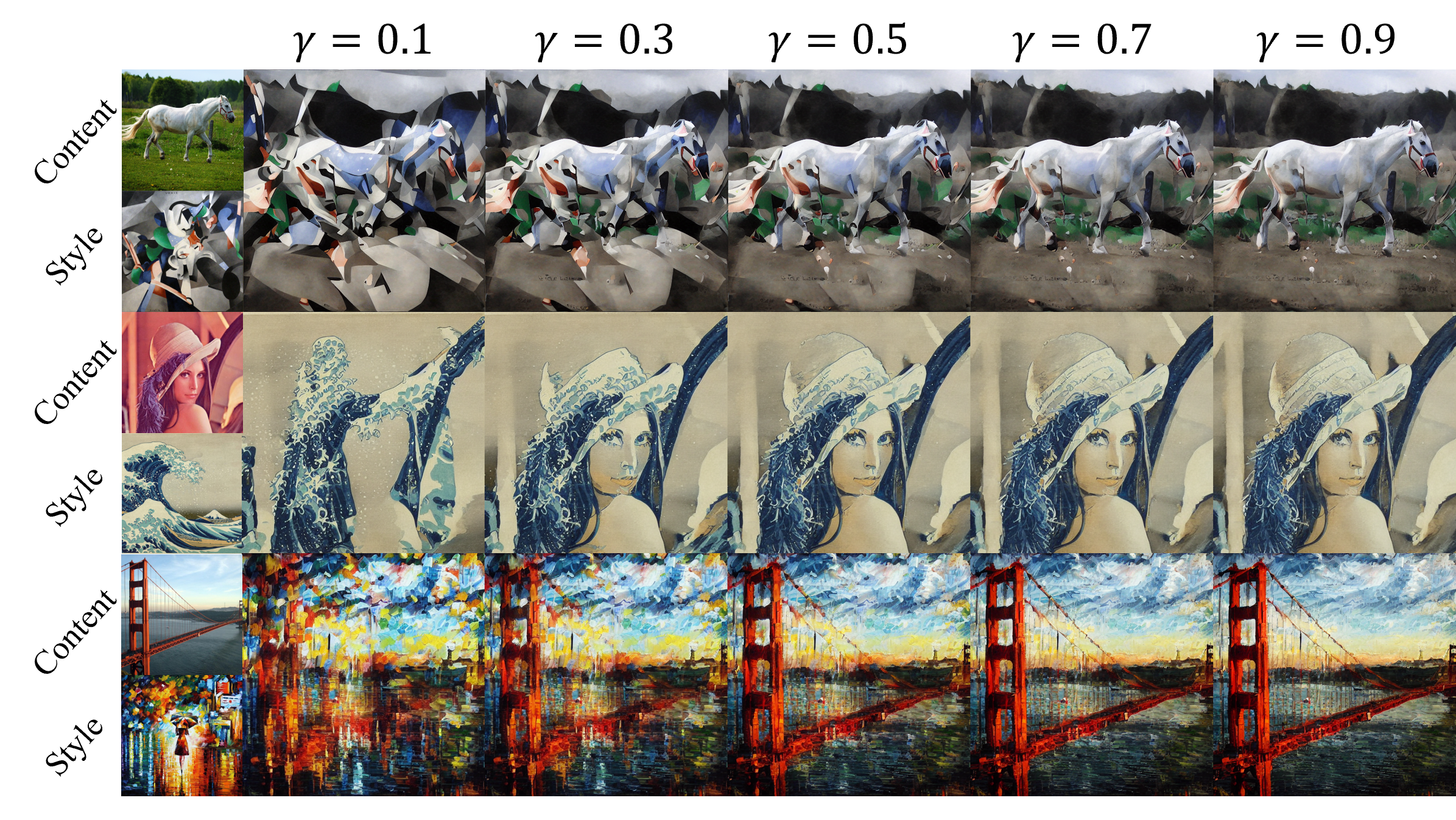}
    \centering
    \vspace{-0.3cm}
    \caption{
        Visualization of effects of query preservation ratio $\gamma$
    }
    \vspace{-0.4cm}
    \label{fig_gamma_preservation_ratio}
\end{figure}

We also visualize the effects of the style-content trade-off by synthesizing images by adjusting $\gamma$.
As shown in Fig.~\ref{fig_gamma_preservation_ratio}, the lower $\gamma$ highly reflects the style while losing the content of the given image, and vice versa.
This characteristic of the proposed method suggests that the users can adjust the degree of style by following their personal preferences.

\vspace{0.2cm}
\noindent\textbf{Study on the value of $\tau$.}
We observe that the gradual increase of $\tau$ enhances the performance of style transfer, although its effects on enhancement become smaller as $\tau$ goes larger, as shown in Fig.~\ref{fig_trade_off_cs}~(b).
This result implies that the attention temperature scaling effectively works with a simple modification of the magnitude of the attention map.

\begin{figure}[t!]
    \hspace*{0.2cm}
    \includegraphics[width=0.95\columnwidth]{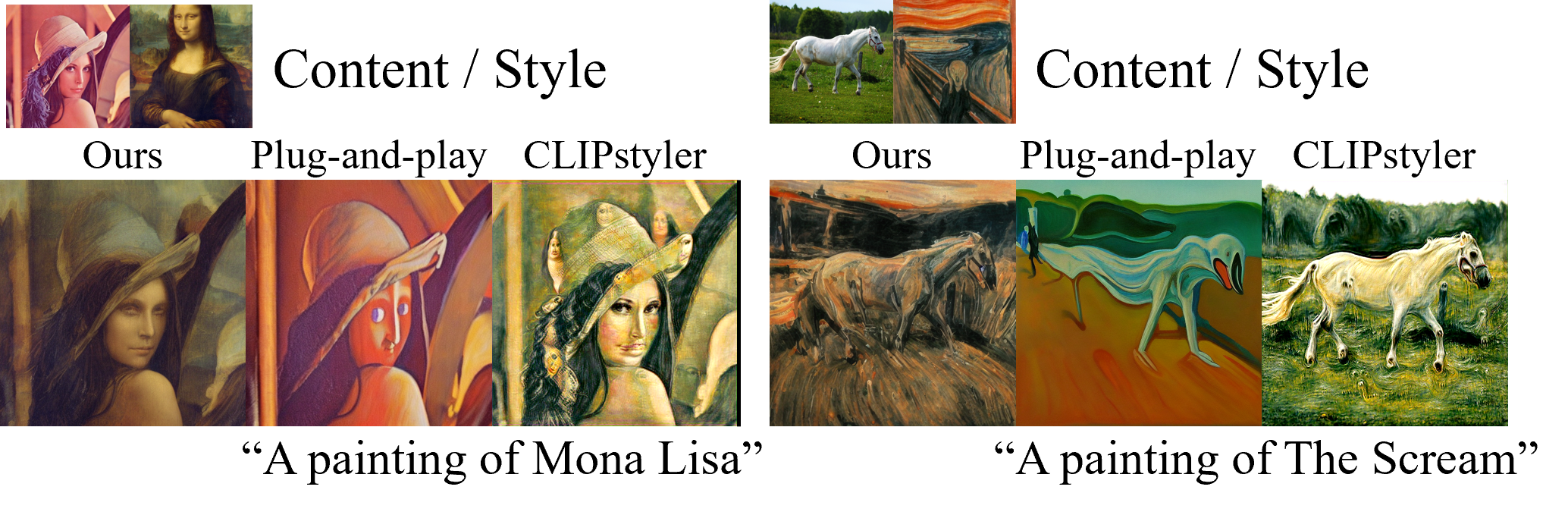}
    \centering
    \vspace{-0.4cm}
    \caption{
        Comparison with text-guided style transfer methods
    }
    \vspace{-0.4cm}
    \label{fig_text_guided}
\end{figure}

\vspace{0.2cm}
\noindent\textbf{Comparison with text-guided style transfer.}
We additionally compare the proposed method with the style transfer methods~\cite{tumanyan2023plug, kwon2022clipstyler} which are conditioned on the textual inputs.
As text-guided methods tend to modify the style largely, we use $\gamma=0.3$ for this experiment.
Since the text condition hardly contains all the information in the style image such as texture and color tones, the transferred results less resemble the target style, as shown in Fig.~\ref{fig_text_guided}.
Differently, we validate that the proposed method successfully transfers the style with high fidelity.

\vspace{-0.1cm}
\section{Conclusion}
\label{sec:conclusion}
Our work addresses the challenges associated with diffusion model-based style transfer methods, which often require time-consuming optimization steps or struggle to leverage the generative potential of large-scale diffusion models. 
To this end, we propose the method of adapting the pre-trained large-scale diffusion model on style transfer in a training-free way.
Our method focuses on manipulating the features of self-attention layers, akin to the cross-attention mechanism, by substituting the \textit{key} and \textit{value} during the content generation process with those of the style. 
Furthermore, we propose the query preservation and attention temperature scaling to mitigate the issue of disruption of content, and initial latent AdaIN to handle the disharmonious color~(failure to transfer the colors of style).
Experimental results show the superiority of our proposed method over state-of-the-art techniques in previous baselines.

\vspace{-0.1cm}
\section*{Acknowledgments}
\vspace{-0.2cm}
This work was supported in part by MSIT/IITP (No. 2022-0-00680, 2019-0-00421, 2020-0-01821, 2021-0-02068), and MSIT\&KNPA/KIPoT (Police Lab 2.0, No. 210121M06).

{
    \small
    \bibliographystyle{ieeenat_fullname}
    \bibliography{main}
}

\clearpage
\setcounter{page}{1}
\maketitlesupplementary
\section{Appendix}

\noindent\textbf{Ablation study for color transfer capability.}
To validate the efficacy of the ablated methods for color transfer, we employ the RGB-$uv$ histogram proposed in HistoGAN~\cite{afifi2021histogan} to measure color transfer capability. Specifically, for a given input image $I$, we convert it into the log-chroma space. For example, choosing the R color channel as the primary and normalizing by G and B yields:
\begin{equation}
\label{eq_uv}
I_{uR}(x)=\log(\frac{I_R(x)+\epsilon}{I_G(x)+\epsilon}), I_{vR}(x)=\log(\frac{I_R(x)+\epsilon}{I_B(x)+\epsilon})
\end{equation}
where the $I_R,I_G,I_B$ refer to the color channels of the image $I$, $\epsilon$ is a small constant for numerical stability, and $x$ is the pixel index.

Then, they compute the intensity $I_y(x)=\sqrt{I^2_R(x)+I^2_G(x)+I^2_B(x)}$ for weighted scaling and differentiable the histogram. The final histogram follows:
\begin{equation}
\label{eq_histo}
\textbf{H}(u,v,c)\propto\Sigma_x k(I_{uc}(x),I_{vc}(x),u,v)I_y(x), 
\end{equation}
where $I_{uG},I_{vG},I_{uB},I_{vB}$ are R and B color channels which projected to the log-chroma space similar to Eq.~\ref{eq_uv}, $c\in\{R,G,B\}$, and $k(\cdot)$ is a inverse-quadratic kernel.

We utilize the Histogram Loss~\cite{afifi2021histogan} as a color similarity metric which measures the Hellinger distance between the histograms of stylized and style images.
\begin{equation}
\label{eq_color_matching}
C(\mathbf{H}_g, \mathbf{H}_t) = \frac{1}{\sqrt{2}}\Vert \mathbf{H}^{\frac{1}{2}}_{cs} - \mathbf{H}^{\frac{1}{2}}_s \Vert_2,
\end{equation}
where $\mathbf{H}_{cs}$ and $\mathbf{H}_s$ are color histograms of stylized and style image, respectively, $\Vert \cdot \Vert$ is the standard Euclidean norm, and $\mathbf{H}^{\frac{1}{2}}$ denotes an element-wise square root.
We adopt the default configuration of HistoGAN~\cite{afifi2021histogan}. For a detailed description of the histogram loss, please refer to the original HistoGAN paper~\cite{afifi2021histogan}.

As a result, we evaluate the efficacy of Initial Latent AdaIN in color tone transfer. 
In Tab.~\ref{tab_histo}, each proposed component contributes to transfer the color tone of the given style image.
Especially, we confirm that the Initial Latent AdaIN prominently affects the for transferring of color tones.

\begin{table}[t!]
\centering
\begin{tabular}{cl|c}
\toprule
& \textbf{Configuration} & Histogram Loss~\cite{afifi2021histogan}~$\downarrow$ \\
\bottomrule
\arch{a} & Ours~\footnotesize{($\gamma=0.75$, default)} & \textbf{0.2804} \\
\arch{b} & -~Style Injection        & 0.4637 \\
\arch{c} & -~Attention Scaling  & 0.3029 \\
\arch{d} & -~Initial Latent AdaIN        & 0.5235 \\
\bottomrule
\end{tabular}
\caption{Ablation study for color transfer capability.}
\label{tab_histo}
\end{table}
% -scaling : 0.3029 -attn fuse: 0.4637

\begin{figure}
    \centering
    \includegraphics[width=1.0\columnwidth]{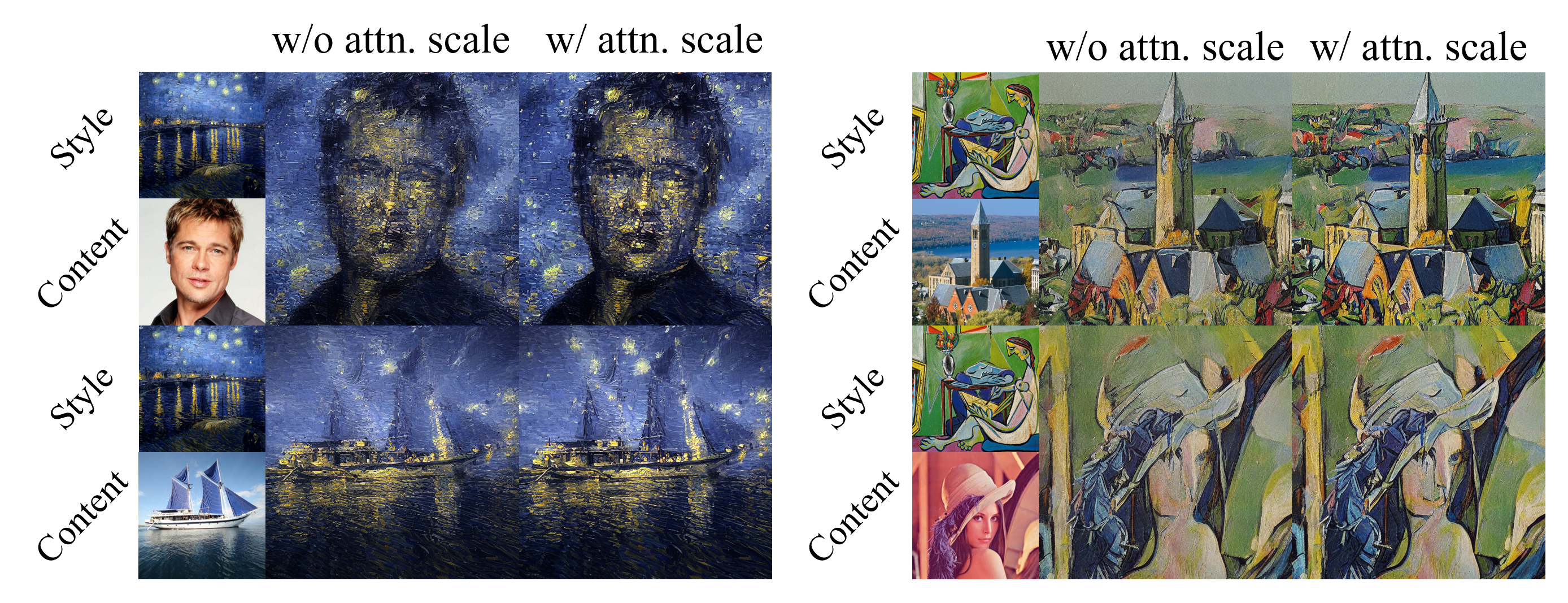}
    \caption{
    Qualitative comparison while ablating the attention temperature scaling. Attention temperature scaling prevents blurry results and helps to keep the local textures in the style image.
    We use $\gamma=0.3$ for this experiment.
    }
    \label{fig:supp_attn_scaling}
\end{figure}

\vspace{0.2cm}
\noindent\textbf{Qualitative comparison with ablation of attention temperature scaling.}
To highlight the effects of attention temperature scaling, we provide some examples of style transfer results while ablating the attention scaling.
As shown in Fig.~\ref{fig:supp_attn_scaling}, we validate that the attention scaling makes the model to synthesize sharp images and well-preserve the patterns in the given style image~(e.g. stars in left example).
This experimental result confirms the significance of the proposed attention temperature scaling method.
Note that, we use $\gamma=0.3$ for this experiment, to keep the strong effect of style transfer in visualization.

\vspace{0.2cm}
\noindent\textbf{Quantitative comparison in the other set.}
In Tab.~\ref{tab_supp_newset}, we conduct quantitative experiments on a new set of style-content pairs~(20 contents, 40 styles) randomly sampled without any overlap with original images.
As reported, the performance enhancement of the proposed method still holds, confirming hyperparameters are well-generalized.

LPIPS is affected by texture and color, as it is based on CNN features~\cite{geirhos2018imagenet}.
To evalute the content and color independently, we measure LPIPS-Grayscale and Histogram-loss in supplementary against the recent and lowest ArtFID baselines~(AesPA-Net, InST, AdaAttN).
As reported in Tab.~\ref{tab_supp_newset}, ours achieves lowest LPIPS-Grayscale, and highest color similarity.

\begin{table}
\centering
\small
\setlength{\tabcolsep}{4pt} % 간격을 조절하려면 이 부분을 조절
\renewcommand{\arraystretch}{0.8} % row 간격
\begin{tabular}{l|cccc}
\toprule
 & Ours & AesPA-Net & AdaAttN & InST \\ 
\midrule
ArtFID  & \textbf{30.38} & 34.55 & 31.87 & 39.11  \\
FID  & \textbf{18.87} & 21.09 & 19.21 & 20.46 \\
LPIPS & \textbf{0.528} & 0.563 & 0.576 & 0.822 \\
LPIPS-Gray & \textbf{0.417} & 0.443 & 0.450 & 0.731 \\
% CFSD    & 0.329 & 0.474 & 0.303 & 0.356 & 0.969 \\
Histogram-loss & \textbf{0.303} & 0.321 & 0.331 & 0.653 \\
% color-histo	0.3033	0.3219	0.331	0.6536
% Ours* 지운다 다 쓰고나면 내가 지워놓음
\bottomrule
\end{tabular}
\vspace{-0.2cm}
\caption{Quantitative comparison in newly sampled test set.}
\vspace{-0.2cm}
\label{tab_supp_newset}
\end{table}

\begin{figure}[t!]
    \vspace{-0.0cm}
    \centering
    \includegraphics[width=0.8\columnwidth]{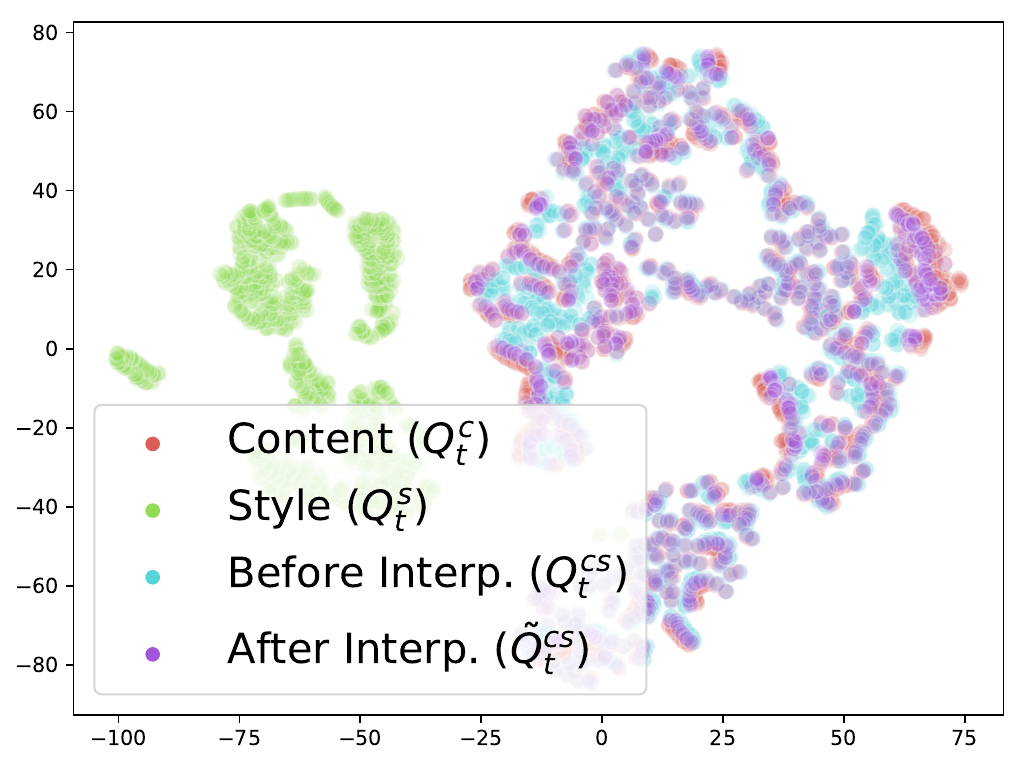}
    \caption{
       t-SNE visualization of query in SA for a style-content pair. 
       Query of content, style, and stylized ones~($Q^{cs}_t$ and $\tilde{Q}^{cs}_t$) at $t$=20 and 7$^{\text{th}}$ decoder layer are used for visualization.
    } 
  \label{fig_tsne_query_preservation}
  \vspace{-0.2cm}
\end{figure}

\vspace{0.2cm}
\noindent\textbf{Analysis on feature space of query preservation.}
Fig.~\ref{fig_tsne_query_preservation} visualizes features of $Q^c_t$, $Q^s_t$, $Q^{cs}_t$, and $\tilde{Q}^{cs}_t$ for a style-content pair. As shown, interpolated features~($\tilde{Q}^{cs}_t$) are located in in-distribution nearby contents, since we gradually combine content query~($Q^c_t$) and stylized one~($Q^{cs}_t$) along with entire reverse process.

Furthermore, we compute the average distance of $\tilde{Q}^{cs}_t$ toward top-5 Nearest Neighbors~(NNs) in (content, style, itself) and the number of them in NNs for all injected layers with $t$=[10, 20, 30, 40].
Distances and \# NNs are (5.49, 9.06, 4.43), (1.24, 0.00, 3.76), implying $\tilde{Q}^{cs}_t$ residing in in-distribution nearby content.

\vspace{0.2cm}
\noindent\textbf{Style transfer with text prompts.}
In this paragraph, We exploit text prompt, obtained by BLIP~\cite{li2022blip}, for DDIM inversion instead of null text token. 
Images in `data\_vis' in the official repository are used, in which easy to caption as they mostly consist of single object.
As a result, ours w/ text shows slight improvement as in Tab.~\ref{tab_blip}.

\vspace{0.2cm}
\noindent\textbf{User study.}
We compare ours with AesPA-Net and InST, the most recent conventional and diffusion methods, for 18 users and 10 examples per user.
We observe that (57.2\%, 76.7\%) of users prefer the proposed method over (AesPA-Net, InST). Note that, ours has a much faster inference speed than InST.

\begin{table}[t!]
\centering
\footnotesize
\setlength{\tabcolsep}{4pt} % 간격을 조절하려면 이 부분을 조절
\begin{tabular}{c|ccc}
\toprule
& ArtFID & FID & LPIPS-Gray \\
\bottomrule
Ours~\footnotesize{(w/ empty prompt, default)} & 34.9 & 21.2 & \textbf{0.47} \\
Ours~\footnotesize{(w/ BLIP prompt)} & \textbf{34.5} & \textbf{20.9} & \textbf{0.47} \\
\bottomrule
\end{tabular}
\caption{Ablation study of the null text token in the diffusion process.}
\label{tab_blip}
\end{table}

\begin{figure}[t!]
    \vspace{0.2cm}
    \hspace*{-0.4cm}
    \includegraphics[width=1.0\columnwidth]{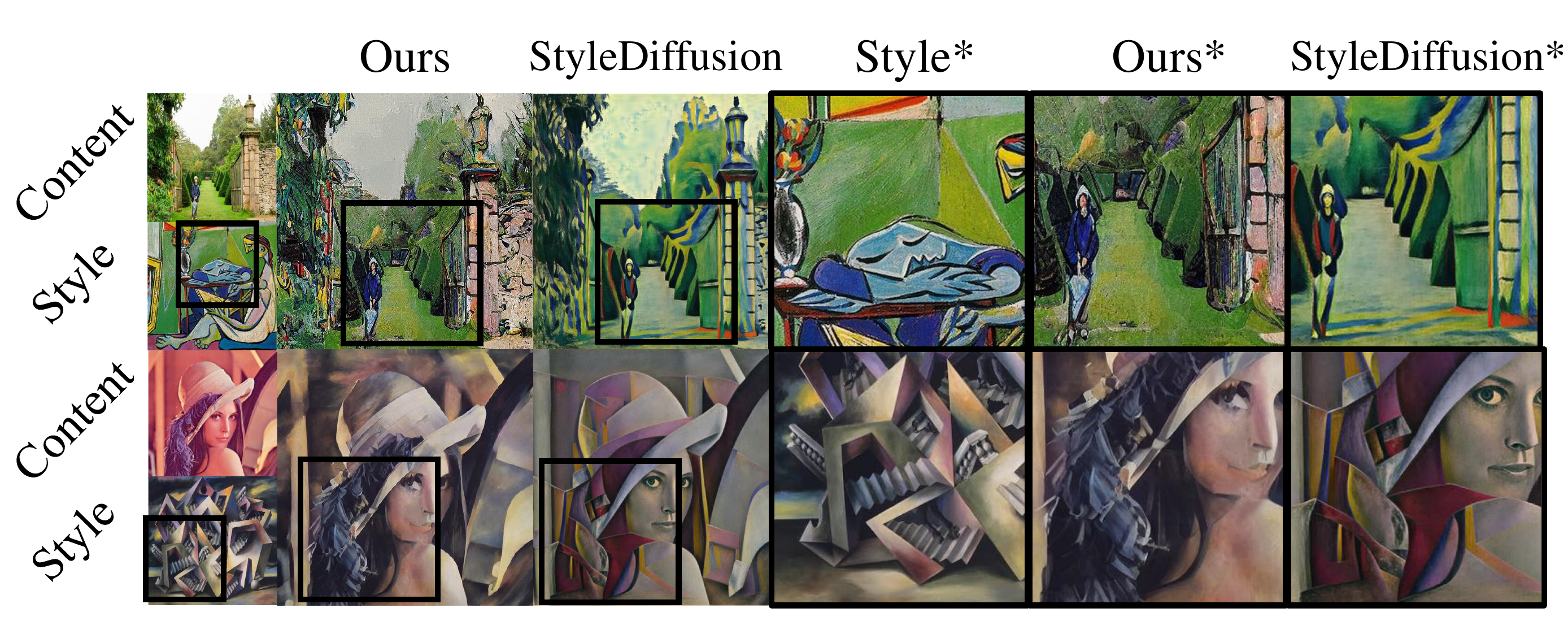}
    \centering
    % \vspace{-0.4cm}
    \caption{
        Qualitative comparison with StyleDiffusion. * denotes cropped version of images. We use $\gamma=0.5$ for visualization.
    }
    \label{fig_supp_comparison_stylediffusion}
\end{figure}

\vspace{0.2cm}
\noindent\textbf{Qualitative comparison with StyleDiffusion.}
As the implementation of StyleDiffusion~\cite{wang2023stylediffusion} is unavailable, we compare ours with examples in supplementary of StyleDiffusion~\cite{wang2023stylediffusion}.
We obtain style-content pairs of StyleDiffusion in repositories of their baselines.
We observe that ours is more suitable for transferring local textures, while StyleDiffusion tends to change the structure of the image significantly, as shown in Fig.~\ref{fig_supp_comparison_stylediffusion}.
We hypothesize that optimizing the style in CLIP~\cite{radford2021learning}'s semantically rich feature space forces StyleDiffusion to be trained in that manner.

\begin{figure}[t!]
    \includegraphics[width=1.0\columnwidth]{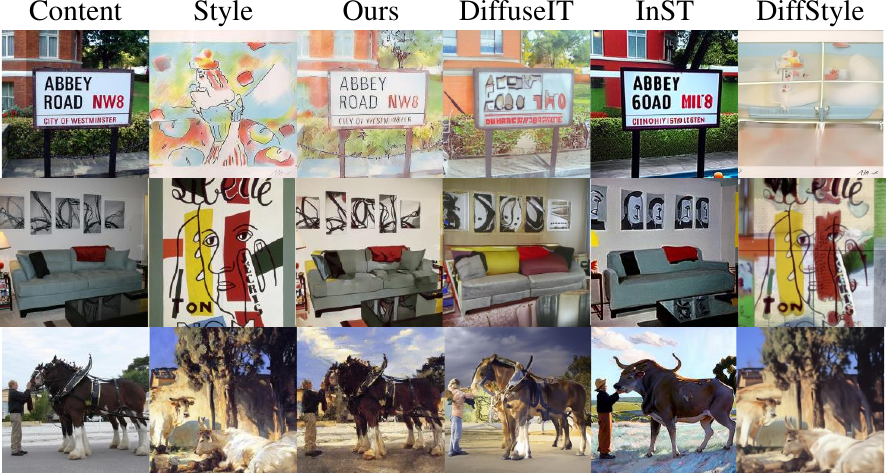}
    \centering
    \vspace{-0.4cm}
    \caption{
    Qualitative comparisons with diffusion-based baselines
    }
    \vspace{-0.5cm}
    \label{fig_supp_diff_qual}
\end{figure}

\begin{figure*}
    \centering
    \includegraphics[width=1.0\textwidth]{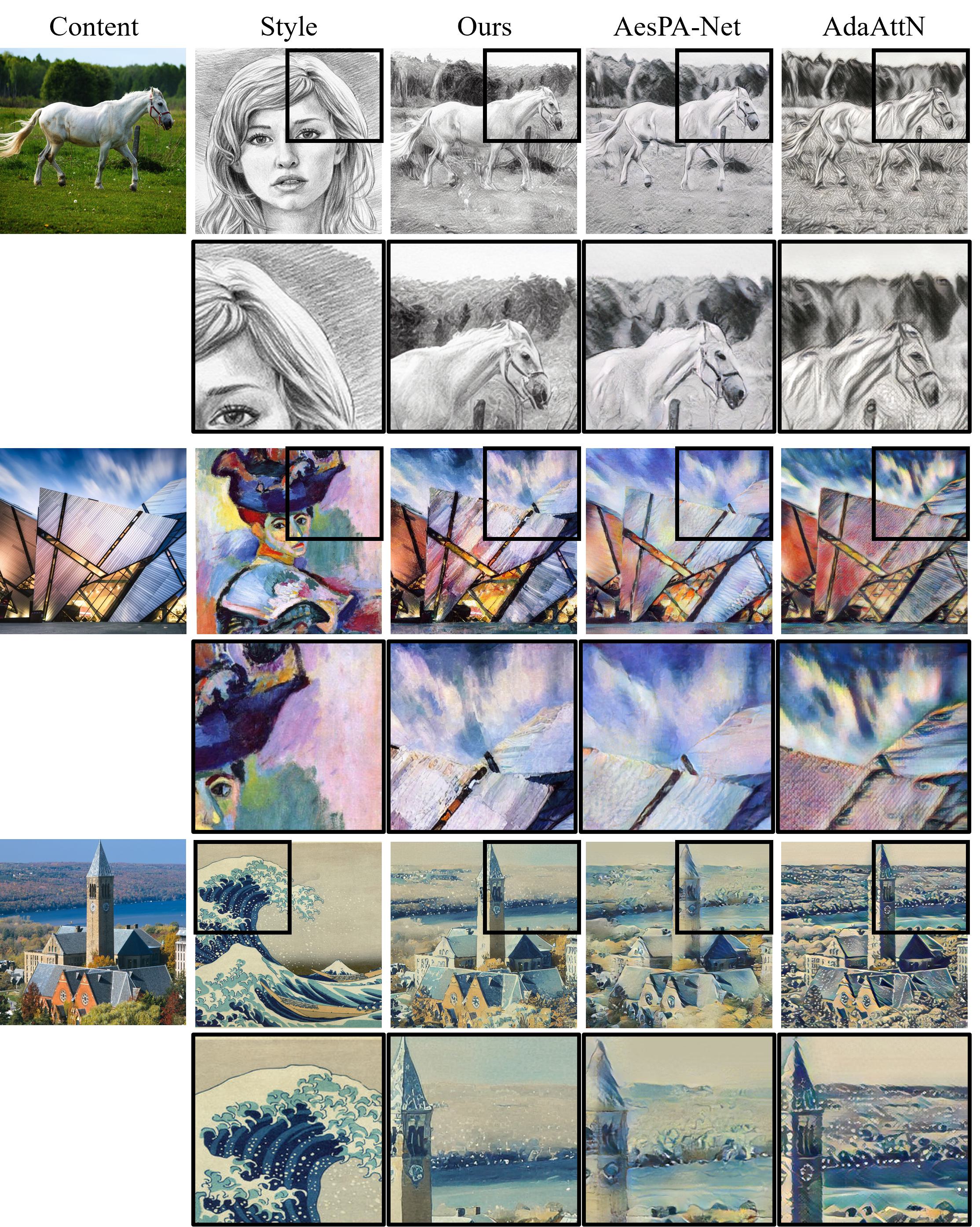}
    \caption{
    Qualitative comparison with baselines~(AesPA-Net, AdaAttN).
    For visualizing the detailed textures, we provide the cropped version of the style image and its stylized counterparts in the second row of every content-style pair. Zoom in for viewing details.
    }
    \label{fig:supp_example_1}
\end{figure*}

\begin{figure*}
    \centering
    \includegraphics[width=1.0\textwidth]{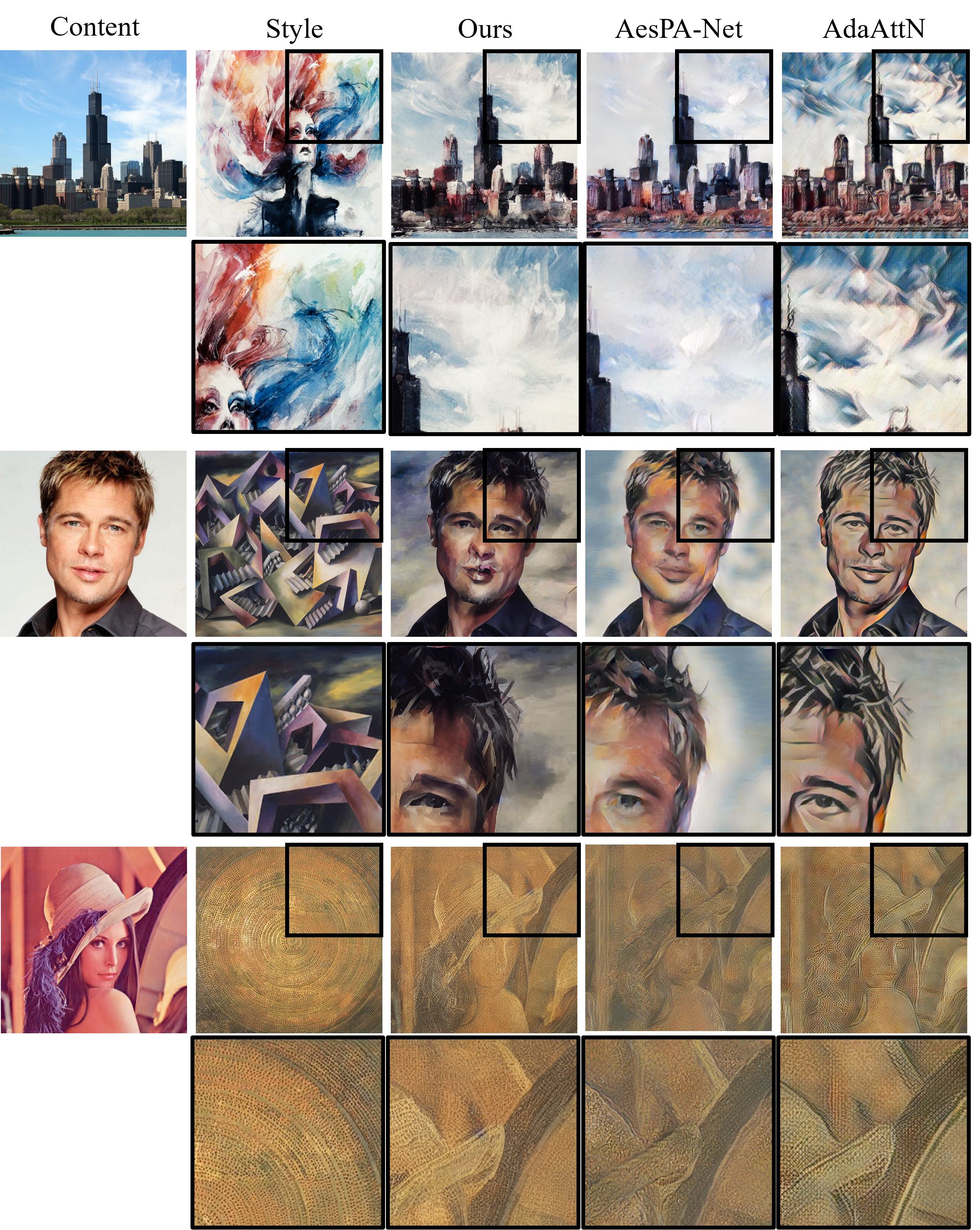}
    \caption{
    Qualitative comparison with baselines~(AesPA-Net, AdaAttN).
    For visualizing the detailed textures, we provide the cropped version of the style image and its stylized counterparts in the second row of every content-style pair. Zoom in for viewing details.
    }
    \label{fig:supp_example_2}
\end{figure*}

\vspace{0.2cm}
\noindent\textbf{Additional qualitative results.}
We additionally compare the proposed method with the most recent baseline~(AesPA-Net) and baseline with the lowest ArtFID~(AdaAttN).
Fig.~\ref{fig_supp_diff_qual} shows the additional qualitative comparison of ours with diffusion model baselines.
Moreover, as shown in Fig.~\ref{fig:supp_example_1},~\ref{fig:supp_example_2}, we observe that ours better-transfers the local texture of a given style into the content image.

\begin{figure*}
    \centering
    \includegraphics[width=1.0\textwidth]{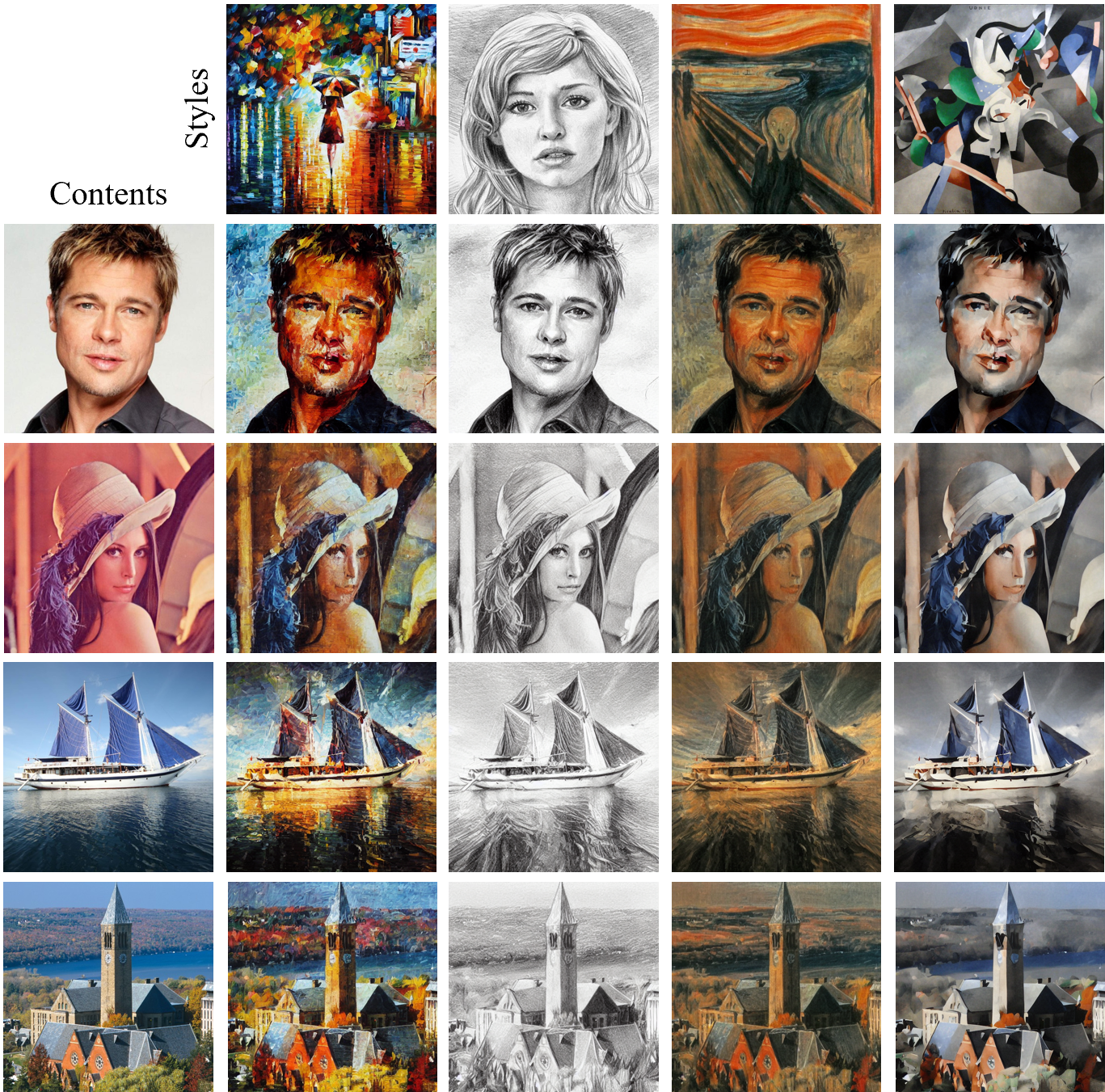}
    \caption{Style transfer results of style and content image pairs. Zoom in for viewing details.}
    \label{fig:supp_example_matrix_1}
\end{figure*}

\begin{figure*}
    \centering
    \includegraphics[width=1.0\textwidth]{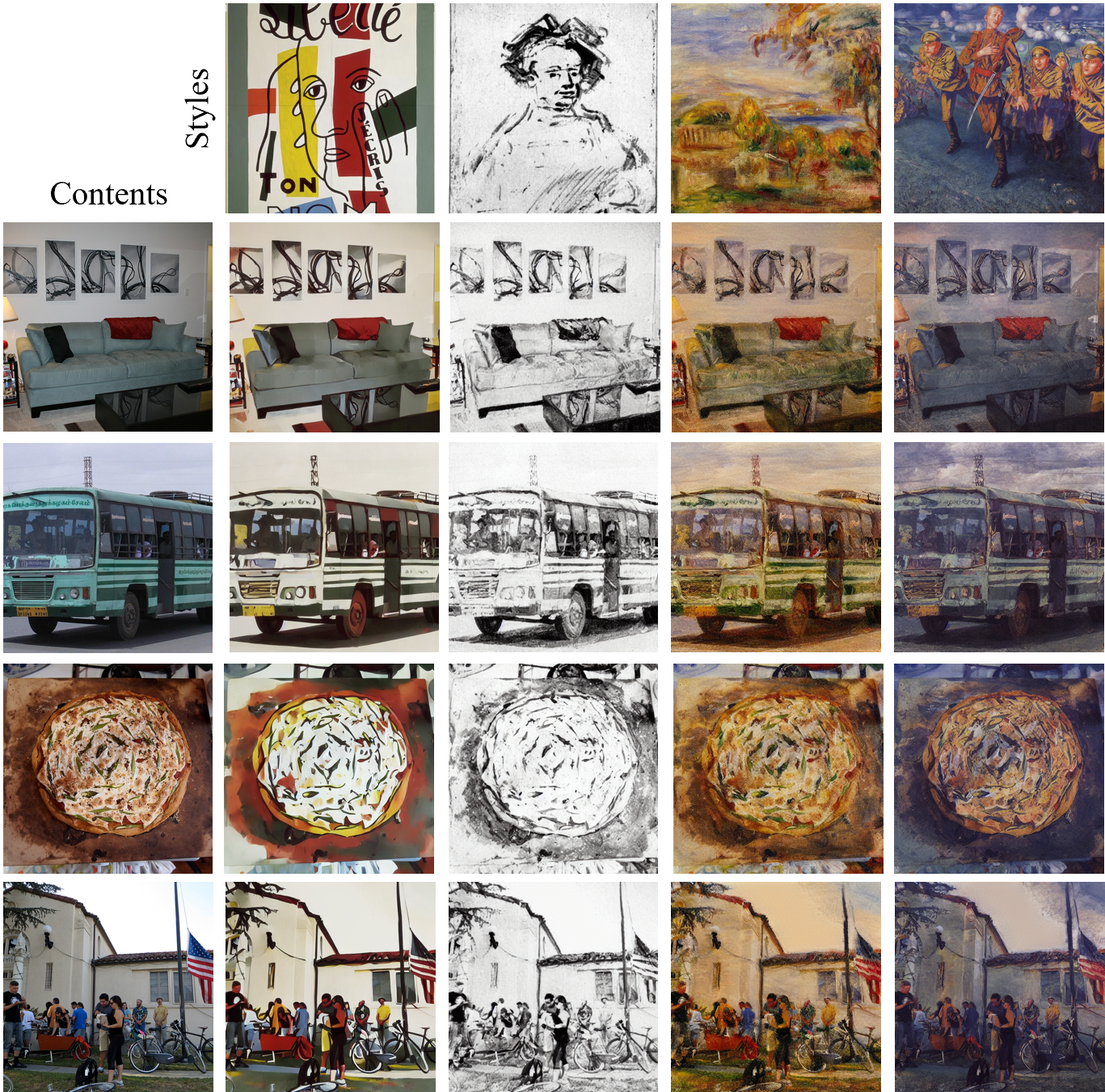}
    \caption{Style transfer results of style and content image pairs. Zoom in for viewing details.}
    \label{fig:supp_example_matrix_2}
\end{figure*}

Also, in Fig.~\ref{fig:supp_example_matrix_1},~\ref{fig:supp_example_matrix_2}, we visualize the style transfer results of various pairs of content and style images.

\end{document}